\documentclass[final]{colt2026}

\usepackage{silence}
\usepackage{times}
\usepackage{hyperref}
\usepackage{amsfonts,bm}
\usepackage{mathrsfs}
\usepackage[format=plain, justification=justified, font=normalsize]{caption}
\usepackage[utf8]{inputenc}				
\usepackage[T1]{fontenc}				
\usepackage{mathtools}
\usepackage{relsize}
\usepackage[dvipsnames]{xcolor}
\usepackage{anyfontsize}
\DeclareMathAlphabet{\mathpzc}{OT1}{pzc}{m}{it}

\graphicspath{{Figures/}}

\hypersetup{
	hidelinks,
	colorlinks=true,
	linkcolor=Blue,
	filecolor=Blue,
	citecolor=Blue,
	urlcolor=black,
  	pdftitle={On the Stability of Nonlinear Dynamics in GD and SGD},
	breaklinks=true
  }

\newcommand{\loss}{f}
\newcommand{\stochasticLoss}{\hat{\mathcal{L}}}
\newcommand{\FullLoss}{\mathcal{L}}
\newcommand{\R}{\mathbb{R}}
\newcommand{\batch}{\mathcal{B}}
\newcommand{\E}{ \mathbb{E} }
\newcommand{\Prob}{ \mathbb{P} }
\newcommand{\N}{ \mathbb{N} }
\newcommand{\C}{ \mathbb{C} }
\newcommand{\ie}{\emph{i.e.}}
\newcommand{\eg}{\emph{e.g.}}
\newcommand{\bmu}{\boldsymbol{\mu}}
\newcommand{\A}{\boldsymbol{A}}

\newcommand{\zeroVec}{\boldsymbol{0}}
\newcommand{\PP}{\boldsymbol{P}}
\newcommand{\DD}{\boldsymbol{D}}
\newcommand{\TT}{\boldsymbol{T}}
\newcommand{\HH}{\boldsymbol{H}}
\newcommand{\MM}{\boldsymbol{M}}
\newcommand{\YY}{\boldsymbol{Y}}
\newcommand{\JJ}{\boldsymbol{J}}
\newcommand{\vv}{\boldsymbol{v}}
\newcommand{\qq}{\boldsymbol{q}}
\newcommand{\pp}{\boldsymbol{p}}
\newcommand{\uu}{\boldsymbol{u}}
\newcommand{\IdentityMat}{\boldsymbol{I}}

\newcommand{\transpose}{\mathrm{T}}
\newcommand{\params}{\boldsymbol{x}}
\newcommand{\param}{x}
\newcommand{\lin}{\mathrm{lin}}
\newcommand{\bPsi}{\boldsymbol{\Psi}}
\newcommand{\D}{\mathcal{D}}
\newcommand{\mD}{\mathlarger{\mathpzc{D}} }
\newcommand{\GDtransform}{\boldsymbol{\psi}}
\newcommand{\SGDtransform}{\hat{\boldsymbol{\psi}}}
\DeclareMathOperator*{\argmax}{arg\,max}

\title[On the Stability of Nonlinear Dynamics in GD and SGD]{On the Stability of Nonlinear Dynamics in GD and SGD:\\Beyond Quadratic Potentials}

\coltauthor{\Name{Rotem Mulayoff} \Email{rotem.mulayoff@cispa.de}\\%
\Name{Sebastian U. Stich} \Email{stich@cispa.de}\\%
\addr{CISPA Helmholtz Center for Information Security, Saarbr\"ucken, Germany}}

\begin{document}

\maketitle

\begin{abstract}
    The dynamical stability of the iterates during training plays a key role in determining the minima obtained by optimization algorithms. For example, stable solutions of gradient descent (GD) correspond to flat minima, which have been associated with favorable features. While prior work often relies on linearization to determine stability, it remains unclear whether linearized dynamics faithfully capture the full nonlinear behavior. Recent work has shown that GD may stably oscillate near a linearly unstable minimum and still converge once the step size decays, indicating that linear analysis can be misleading. In this work, we explicitly study the effect of nonlinear terms. Specifically, we derive an exact criterion for stable oscillations of GD near minima in the multivariate setting. Our condition depends on high-order derivatives, generalizing existing results. Extending the analysis to stochastic gradient descent (SGD), we show that nonlinear dynamics can diverge in expectation even if a single batch is unstable. This implies that stability can be dictated by a single batch that oscillates unstably, rather than an average effect, as linear analysis suggests. Finally, we prove that if all batches are linearly stable, the nonlinear dynamics of SGD are stable in expectation.
\end{abstract}

\section{Introduction}\label{Sec:Introduction}
Understanding the nature of the minima reached by our training procedures is a central problem in machine learning and optimization \citep{neyshabur2014search}. A common way to study this issue is by analyzing the stability of the iterates as the algorithm approaches a minimum \citep{wu2018sgd}. For example, it has been shown that stable minimizers of gradient descent (GD) correspond to flat minima \citep{cohen2021gradient}, which have been associated with flat predictor functions \citep{mulayoffNeurips,nacson2023the} and balanced networks \citep{mulayoff2020unique}. These highlight the role of dynamical stability in shaping the properties of the obtained solutions.

In dynamical systems theory, stability analysis is often carried out via linearization. Once the iterates enter a neighborhood of a fixed point, it is often sufficient to study the linearized system in order to determine whether convergence occurs \citep{thompson2002nonlinear}. This technique has been widely applied to study GD \citep{cohen2021gradient}. Extending it to the stochastic regime, \citet{wu2018sgd} proposed using linearization to analyze the stability of stochastic gradient descent (SGD) in the mean-square sense.

This approach has inspired a large body of subsequent research. In particular, \citet{ma2021on} demonstrated that the moments of the linearized dynamics evolve independently, and for the second moment (mean squared error), they provided an implicit expression for the exact stability criterion. Building on this result, \citet{mulayoff24a} derived an explicit form of the condition, yielding new insights into the linear stability of SGD. Importantly, the stability threshold on the step size depends on the curvature of all samples in the training set (see App.~\ref{app:Background on linear stability}). However, despite this progress, it remains unclear whether, and under what conditions, the behavior of linearized iterates truly reflects the full nonlinear dynamics of SGD.

In this work, we address this gap by explicitly examining the role of nonlinear terms. We begin with the deterministic setting of GD, where linearized dynamics fail to predict whether the iterates stay in the vicinity of minimizers. In particular, it has been shown that the iterates may exhibit stable oscillations near a linearly unstable minimum and, after step-size decay, eventually converge to it \citep{chen2023beyond}. These oscillations arise from a flip (period-doubling) bifurcation of the GD map at the edge of stability (EoS), causing the iterates to alternate along the sharpest direction of the minimum. The stability of this bifurcation is determined by the leading nonlinear coefficient in its normal form (see Sec.~\ref{sec:flip bifurcation}). Building on a normal-form analysis of the flip bifurcation, we derive a criterion that characterizes the existence of stable oscillations in the EoS regime. The resulting condition (Thm.~\ref{Thm:Stable oscillations of GD}) depends on the second, third, and fourth order derivatives of the loss at the minimum and both generalizes and corrects existing work.

We then extend our analysis to the stochastic setting of SGD. Following prior work, we focus on interpolating minima and assume the loss functions are analytic in a neighborhood of the minimum. In this setting, linearized dynamics, combined with mean-square analysis, suggest that the stability threshold of SGD depends on an average curvature over the different mini-batches. In contrast, we show that if the iterates oscillate unstably even with respect to a single batch, the full nonlinear dynamics of SGD can be unstable in expectation. This suggests that stability can be determined by one batch, contradicting prior results about curvature averaging over all batches (see Thm.~\ref{Thm:Unstable oscillations in SGD}).

Finally, we provide a sufficient condition for the stability of SGD. Specifically, we prove that if the dynamics are linearly stable with respect to all possible batches, then there exists a neighborhood of the minimum from which the full nonlinear dynamics converge in expectation (see Thm.~\ref{Thm:Sufficient condition for SGD}). Our analysis uses Koopman theory \citep{koopman1931hamiltonian}, which allows us to formulate the finite-dimensional nonlinear dynamics as a linear dynamical system in an infinite-dimensional Hilbert space. This reformulation yields two notable benefits. First, nonlinear dynamics are reduced to linear ones, which are significantly more tractable. Second, the transformation provides a deterministic linear relationship between the moments of the dynamics. Then, we use tools from functional analysis to derive the result. Considering our earlier findings, we see that this sufficient condition can also be necessary in certain cases, as we demonstrate in Sec.~\ref{Warmup:Stability threshold of stochastic gradient descent}. Specifically, this occurs when unstable oscillations arise in batches with the lowest stability threshold.

\section{Preliminaries}\label{sec:Dynamical systems preliminaries}
In this section, we provide an overview of the key definitions and concepts in dynamical systems that are relevant to this work. Consider the discrete-time dynamical system $\params_{t+1} = \GDtransform(\params_t)$, where $\GDtransform:\mathbb{R}^d \to \mathbb{R}^d$ is a differentiable map. We denote by $\GDtransform^k$ the $k$-fold iterate of $\GDtransform$.
\begin{definition}[Stability]\label{def:Stability}
    A fixed point $\params^*=\GDtransform(\params^*)$ is called (locally asymptotically) stable if:
    \begin{enumerate}
        \item For any $\varepsilon>0$ there exists $\delta>0$ s.t.\ if $\|\params-\params^*\|<\delta$, then $ \|\GDtransform^t(\params)-\params^*\|<\varepsilon \  \ \forall t \in\mathbb{N}$.
        \item There exists $\delta>0$ such that if $ \ \|\params-\params^*\|<\delta $, then $ \lim_{t\to\infty} \|\GDtransform^t(\params)-\params^*\|=0 $.
    \end{enumerate}
\end{definition}
A standard approach to studying the stability of a fixed point is to analyze the linearized dynamics around that point. Let $\D \GDtransform$ be the Jacobian of $\GDtransform$, \emph{i.e.}, $ \D \GDtransform(\params) = \frac{\partial \GDtransform}{\partial \params}(\params) $, and $ r(\cdot) $ be the spectral radius, defined as $r(\A) = \max\{ |\lambda| \colon \lambda \in \sigma(\A) \}$, where $\sigma(\A)$ is the set of eigenvalues of~$\A$.
\begin{definition}[Linear Stability]\label{def:Linear stability}
    A fixed point $\params^*$ of $\GDtransform$ is said to be linearly stable if $r \big(\D \GDtransform(\params^*) \big) \! \! < \! 1$. If $r \big(\D \GDtransform(\params^*)\big) \! > \! 1$, then $\params^*$ is said to be linearly unstable.
\end{definition}
A classical result states that linear stability implies local asymptotic stability  \citep{kuznetsov1998elements}. Conversely, if the spectral radius of $ \D \GDtransform(\params^*) $ is strictly larger than one, then $\params^*$ is unstable. The case $r \big(\D \GDtransform(\params^*)\big)=1$ is inconclusive: higher-order terms are needed to determine stability.

The notion of a fixed point naturally extends to periodic points and cycles. Such periodic behavior may arise, for example, in the vicinity of unstable fixed points.
\begin{definition}[Period-$\boldsymbol{k}$ Point]
    A point $\params^*\in\mathbb{R}^d$ is said to be a period-$k$ point if $\GDtransform^k(\params^*) = \params^*$. If $k$ is the smallest positive integer satisfying this property, then $\params^*$ is said to have prime period $k$.
\end{definition}
Every periodic point is associated with a finite orbit, consisting of the distinct points visited under repeated applications of the map.
\begin{definition}[Period-$\boldsymbol{k}$ Cycle]
    A set of $k$ distinct points $\{ \params^{(i)} \}_{i=1}^k$ is said to be a period-$k$ cycle if $ \params^{(i)} = \GDtransform^{i-1}\big(\params^{(1)} \big) \ $ for $ i=2,\ldots,k$ and $\params^{(1)}$ has prime period $k$.
\end{definition}
Under this definition, every point in a period-$k$ cycle is itself a period-$k$ point. As with fixed points, one can study the stability of a cycle through the iterated map $\GDtransform^k$.
\begin{definition}[Stable Period-$\boldsymbol{k}$ Cycle]
    A period-$k$ cycle of $\GDtransform$ is said to be stable if its points are (locally asymptotically) stable fixed points of the map $\GDtransform^k$.
\end{definition}
The stability of a period-$k$ cycle $\big\{ \params^{(i)} \big\}_{i=1}^k$ can be characterized through the linearization of the iterated map. Let $\D \GDtransform^k(\params)$ denote the Jacobian of $ \GDtransform^k$ evaluated at $\params$. By the chain rule,
\begin{equation}
    \D \GDtransform^k \big( \params^{(i)} \big)
    = \frac{\partial \GDtransform^k}{\partial \params}\big( \params^{(i)} \big)
    \sim {\prod\nolimits_{j=1}^{k}}
    \D \GDtransform \big( \params^{(j)} \big) \triangleq \JJ^*,
\end{equation}
where $ \sim $ denotes matrix similarity. Here, we used the fact that the Jacobians corresponding to different points on the cycle differ only by cyclic permutations of the factors in the product, and are therefore similar matrices, with identical spectra. Thus, this analysis is independent of the chosen point on the cycle, and linear stability of one point on the cycle implies linear stability of the entire cycle. Consequently, a cycle is linearly stable if $r\big( \JJ^* \big) < 1$, and linearly unstable if $r \big( \JJ^* \big)>1$.  Intuitively, if the iterates $\params_t$ happen to be sufficiently close to a point belonging to a stable cycle, then they converge asymptotically to that cycle. In contrast, trajectories that arrive near an unstable cycle eventually leave its neighborhood. In this paper, we apply these concepts to the gradient descent map and use them to characterize its behavior in the vicinity of minimizers.

\section{Warmup}\label{Sec:Background}
In this section, we illustrate how nonlinear terms in the dynamics influence the behavior of gradient-based methods. We begin with GD, showing how the sign of the third-order term can control the stability of oscillations. We then turn to SGD, where we find that, contrary to linear predictions, stability in expectation can be dictated by a single batch rather than an average.

\subsection{Normal Form of Oscillations in Gradient Descent}\label{Sec:Warmup oscillations in gradient descent}
As discussed in Sec.~\ref{sec:Dynamical systems preliminaries}, a linearly unstable fixed point is, in particular, asymptotically unstable. While convergence may occur for a measure-zero set of initializations in its vicinity\footnote{More precisely, for initializations lying on the stable manifold of the fixed point.}, generic nearby starting points result in trajectories that eventually leave any sufficiently small neighborhood. Consequently, convergence to a linearly unstable fixed point is highly unlikely to happen in practice.

In the context of GD, for a twice differentiable objective $\FullLoss \colon \R^d \to \R$, a minimizer $\params^*$ corresponds to a linearly unstable fixed point if the step size $ \eta$ is greater than $ \eta_{\lin} = 2/\lambda_{\max}(\nabla^2 \FullLoss(\params^*)) $, where $ \lambda_{\max} $ denotes the top eigenvalue. Here, $ \eta_{\lin} $ is the linear stability threshold on the step size. Recently, it has been shown that GD typically operates at the edge of stability when optimizing neural networks \citep{cohen2021gradient}. In this regime, the largest Hessian eigenvalue along the trajectory remains close to the critical value $2/\eta$ and marginally exceeds it as the iterates approach a minimum. This implies that GD often encounters linearly unstable minima. As we discussed above, convergence to these minimizers is unlikely. Nevertheless, \citet{chen2023beyond} demonstrated that GD may still exhibit stable oscillations in their vicinity and, following a subsequent reduction of the step size, eventually converge to them. Thus, understanding when such oscillations are stable is essential for characterizing the minima that GD can ultimately reach.

\begin{figure}[t]
    \subfigure[Graphs of $ \loss_{+}$ and $ f_{-} $][t]{%
        \includegraphics[width=0.33\linewidth]{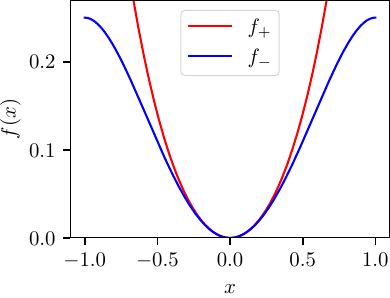}
        \label{fig:Warmup_functions}
    }%
    \subfigure[Bifurcation diagram of $ \loss_{+} $][t]{%
        \includegraphics[width=0.33\linewidth]{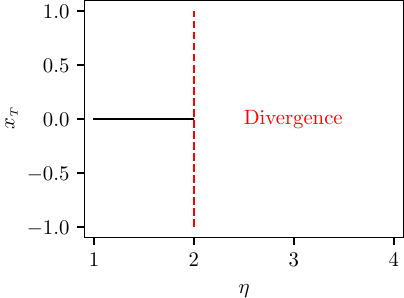}
        \label{fig:Warmup_bifurcation1}
    }%
    \subfigure[Bifurcation diagram of $ \loss_{-} $][t]{%
        \includegraphics[width=0.33\linewidth]{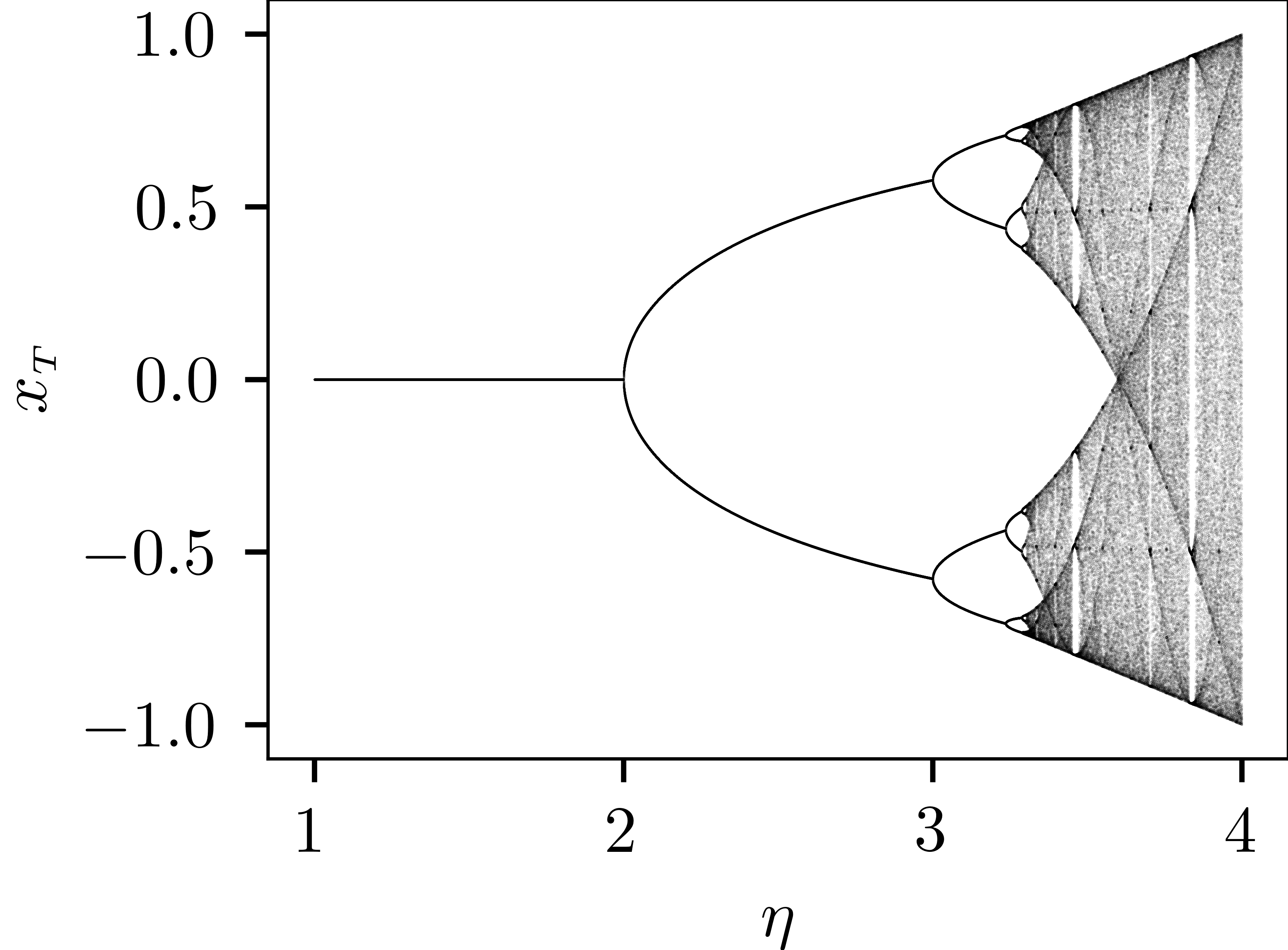}
        \label{fig:Warmup_bifurcation2}
    }
    \caption{{\bfseries Stable vs. unstable oscillations near a minimum.} We apply GD to $ f_{+} $ and~$ f_{-} $ from \protect\eqref{eq:normal forms of flip bifurcation} with various step sizes $\eta \in (1,4)$. The resulting dynamics \protect\eqref{eq:normal form in GD dynamics} correspond to the normal form of a flip bifurcation. Once the step size exceeds the linear stability threshold $ \eta_{\lin} = 2 $, stability is determined by the sign of the cubic term in the dynamics.
    Panel~\protect\subfigref{fig:Warmup_functions} shows $ f_{+} $ and $ f_{-} $, whose minima share the same sharpness.
    Panel~\protect\subfigref{fig:Warmup_bifurcation1} visualizes GD's output on $ f_{+} $ with various step sizes. When the step size $ \eta $ crosses $ \eta_{\lin} $, the minimum $ x^* = 0 $ loses stability, resulting in unstable oscillations, which lead to divergence.
    Panel~\protect\subfigref{fig:Warmup_bifurcation2} depicts GD's convergent points on $ f_{-} $ for various step sizes. At the threshold,  $ \eta = \eta_{\lin} $, the minimizer $ x^*=0 $ loses stability, and the iterates settle into a stable period-$2$ cycle, which then undergoes period doubling, chaos, and eventually divergence for $\eta > 4$.}
\end{figure}

In this section, we demonstrate how nonlinear terms in the dynamics influence the behavior of GD beyond $\eta_{\lin}$. Let us consider GD's iterates over two univariate functions, depicted in  Fig.~\ref{fig:Warmup_functions}, that share the same curvature at the minimum but differ in higher-order terms:
\begin{equation}\label{eq:normal forms of flip bifurcation}
    \loss_{+}(\param) = \frac{1}{2}\param^2+\frac{1}{4}\param^4, \qquad \text{and} \qquad \loss_{-}(\param) = \frac{1}{2}\param^2-\frac{1}{4}\param^4.
\end{equation}
Both have the same sharpness at the local minimizer $\param^* = 0$, with $ \loss_{+}''(0) = \loss_{-}''(0) = 1 $, yielding a linear stability threshold of $ \eta_{\lin} = 2 $. The iterates of GD are given by
\begin{equation}\label{eq:normal form in GD dynamics}
    \param_{t+1} = -(\eta-1)\param_{t} \pm \eta \param_t^3 .
\end{equation}
Let us examine how the asymptotic value of the iterates depends on the step size~$ \eta $. Figures~\ref{fig:Warmup_bifurcation1} and~\ref{fig:Warmup_bifurcation2} plot the accumulation points of $ \{ \param_t \}$ for various values of $\eta$ on $\loss_{+}$ and $\loss_{-}$, where $ \param_0 $ is chosen at random from the interval $ (-1,1)$. For $ \eta < \eta_{\lin} $, both dynamics converge to $ \param^* = 0 $. However, when $ \eta >  \eta_{\lin} $, the behavior of the dynamics differs. The iterates on $\loss_{+}$ immediately diverge once the step size crosses $\eta_{\lin}$. In contrast, GD on $ \loss_{-} $ exhibits rich dynamics, where it initially settles into stable cycles while undergoing a sequence of period-doubling bifurcations over a wide range of step sizes, before transitioning into chaos, and finally diverging once $ \eta  >  4 $. Importantly, when such stable oscillations occur, decaying the step size below $\eta_{\lin}$ results in convergence to $\param^*$.

This simple example demonstrates that exceeding the linear stability threshold does not necessarily imply that GD escapes the vicinity of a minimum. Interestingly, under mild assumptions, the behavior of any nonlinear dynamics along the critical manifold near a linearly unstable fixed point can be reduced to this simple one-dimensional map, called \emph{normal form} (see Sec.~\ref{sec:flip bifurcation}). Then, as the example illustrates, the sign of the cubic term in this normal form can be used to determine whether stable oscillations arise. In Sec.~\ref{Sec:Oscillations in gradient descent} we extend this analysis to higher dimensions and derive a general condition for stable oscillations of GD at the edge of stability.

\subsection{Stability of Stochastic Gradient Descent Governed by a Single Batch}\label{Warmup:Stability threshold of stochastic gradient descent}
Linearized analyses of SGD in expectation predict stability by averaging curvature information across all samples (see App.~\ref{app:Background on linear stability}). In particular, under mean-square analysis, the expected squared distance of the iterates from a minimizer remains bounded as long as the step size $\eta$ is below a threshold that depends on the curvature of all samples through an averaging effect. Here, we show that the actual nonlinear dynamics of SGD can behave fundamentally differently. In certain cases, stability in expectation is governed by a single batch rather than by an averaged stability criterion.

To illustrate this discrepancy, we examine the dynamics of SGD on the following functions:
\begin{equation}\label{eq:counter example}
    \loss_+(\param) = \frac{1}{2}\param^2 + \frac{1}{4}\param^4, \qquad
    \text{and} \qquad
    \loss_a(\param) = \frac{a}{2}\param^2,
\end{equation}
where $ a \in (0,1) $ is a fixed parameter. More specifically, we consider the minimization of the average of~$ \loss_+ $ and~$ \loss_a $, where at each iteration, SGD takes a gradient step with respect to one of these functions, chosen at random. Here $ \param^* = 0 $ is an interpolating minimizer, \ie, it minimizes each function individually. The sharpness of these functions at $ \param^* $, given by their second derivative, is $ h_+ = \loss_+''(0) = 1 $ and $ h_a = \loss_a''(0) = a $. Consequently, the linear stability thresholds for optimizing each function separately are $ \eta_+ = 2/h_+= 2 $ and $\eta_a = 2/h_a = 2/a $. Under the linearized mean-square analysis of SGD, the combined stability threshold equals (see App.~\ref{app:Background on linear stability})
\begin{equation}
    \eta_{\lin} = 2\frac{h_+ + h_a}{h_+^2 + h_a^2} = 2\frac{1+a}{1+a^2} > 2.
\end{equation}

We now compare this prediction with the actual nonlinear dynamics. Despite the linearized analysis predicting stability for all $\eta < \eta_{\lin}$, the following proposition shows that the SGD iterates diverge in expectation whenever $\eta>2$ (see proof in App.~\ref{Appendix::Worst case condition proof}).
\begin{proposition}[Worst-Case Batch]\label{prop:Worst case condition}
    Let $ \{ \param_t \} $ be SGD's iterates on $\loss_+$ and $\loss_a$ from \eqref{eq:counter example}, s.t. $ \param_0 \neq 0 $. If $\ \eta > 2$ then $ \E\big[ |\param_t-\param^*| \big] \underset{t \to \infty}{\longrightarrow} \infty$. 
\end{proposition}
In other words, because one of the two losses ($\loss_+$) becomes unstable at $\eta > 2$, the entire stochastic process diverges despite the linearized analysis predicting stability up to $\eta_{\lin}>2$.
This simple example shows that nonlinear SGD can be governed by the least stable batch rather than by an average stability criterion. In Sec.~\ref{Sec:Stability of nonlinear dynamics in SGD}, we formalize this observation and provide general necessary and sufficient conditions for nonlinear stability of SGD.

\section{Oscillations in Gradient Descent at the Edge of Stability}\label{Sec:Oscillations in gradient descent}
In this section, we present a general condition for stable oscillations of GD near minima and discuss its relation to prior work. As noted earlier, GD typically exhibits the EoS phenomenon when optimizing neural networks \citep{cohen2021gradient}. During the early stages of training, a phase called \emph{progressive sharpening}, the landscape becomes sharper as the top eigenvalue of the Hessian increases until it reaches the linear stability threshold of $ 2/\eta $ \citep{wang2022analyzing}. Beyond this point, the sharpness remains slightly above $ 2/\eta $ for the rest of the training. Consequently, as the iterates approach a minimum, they often encounter minima whose sharpness marginally exceeds the linear stability threshold. While convergence to such minimizers is highly unlikely in practice (see Sec.~\ref{Sec:Warmup oscillations in gradient descent}), the iterates can stably oscillate in their vicinity. Then, once the step size decays, these oscillations vanish, allowing the method to settle into the minimum. Thus, understanding GD behavior at the EoS near minima is critical for determining to which minima it converges.

Recently, \citet{chen2023beyond} studied this problem and, in the univariate setting, derived the exact condition for the existence of stable oscillations at the edge of stability, which correspond to period-$2$ cycles. Specifically, let $\FullLoss \colon \mathbb{R} \to \R$ be four times differentiable in a neighborhood of a local minimizer $\param^*$. Then they showed that stable period-$2$ cycles exist when 
\begin{equation}\label{eq:GD univariate exact stability condition}
    {3 \frac{ \big( \FullLoss''' (\param^*)\big) ^2} { \FullLoss''(\param^*) } >  \FullLoss''''(\param^*).}
\end{equation}
Yet, it is quite difficult to extrapolate the exact condition from this univariate result to higher dimensions. For the multivariate case $ \FullLoss \colon \R^d \to  \R $, they suggested that this condition should serve as a necessary condition when evaluated along the sharpest direction of the Hessian. Formally, let~$ \vv_{\max} $ be an eigenvector corresponding to the largest eigenvalue of $ \nabla^2 \FullLoss $ at a minimizer $ \params^* $, then their hypothesized necessary condition is:
\begin{equation}\label{eq:GD sufficient condition}
    {3 \frac{ \big(  \D^3 \FullLoss  ( \params^* ) [\vv_{\max} ]^3 \big) ^2 }{ \D^2 \FullLoss  ( \params^* ) [\vv_{\max} ]^2  }> \D^4  \FullLoss ( \params^* )[\vv_{\max} ]^{4},}
\end{equation}
where $ \D^k $ denotes the $k$th derivative in multilinear form\footnote{$ \D^k \FullLoss( \params ) \big[ \vv \big]^p \big[ \qq \big]^m $ is defined as the $ k $th derivative of $ \FullLoss $ at $ \params $, applied $ p $ times to $  \vv $ and $ m $ times to $ \qq $ ($ p + m \leq k $).}.  
However, this hypothesis turns out to be incorrect, as we show below. In particular, the suggested condition above fails to generalize the exact univariate condition \eqref{eq:GD univariate exact stability condition} to the multivariate setting. Here we present the exact condition for stable oscillations in the multivariate case, which takes the following form.
\begin{theorem}[Stable Oscillations in GD at the Edge of Stability]\label{Thm:Stable oscillations of GD}
    Let $ \FullLoss \colon \R^d \to \R $ and $ \params^* $ be its local minimizer,  such that $ \FullLoss $ is four times differentiable at $ \params^* $. Assume $ \nabla^2 \FullLoss(\params^*)$ is positive definite and let $ \vv_{\max} $ be a top eigenvector corresponding to the maximal eigenvalue. Suppose that the step size is taken sufficiently close to the linear stability threshold from above, i.e., $ {\eta} \to \eta_{\lin}^+$ with $ \eta_{\lin} = {2}/ \lambda_{\max} \big(\nabla^2 \FullLoss( \params^* )  \big)$, and that $ \lambda_{\max} $ has multiplicity one. Then a stable period-$2$ cycle exists in the vicinity of $ \params^* $ if and only if
    \begin{equation}\label{eq:condition for stable oscillations}
        \D^3 \FullLoss( \params^* ) \big[ \vv_{\max} \big]^2 \big[ \qq \big] > \D^4 \FullLoss( \params^* ) \big[ \vv_{\max} \big]^{4},
    \end{equation}
    where
    \begin{equation}
        \qq \triangleq \Big[ \nabla^2 \FullLoss( \params^* )  \Big]^{-1} \nabla_{\vv} \D^{3} \FullLoss( \params^* ) \big[ \vv \big]^3  \Big|_{\vv = \vv_{\max}}.
    \end{equation}
\end{theorem}
This theorem states that GD can stably oscillate near a minimum if and only if the condition in~\eqref{eq:condition for stable oscillations} holds. This condition is composed of high-order derivatives of the loss. Intuitively, it suggests that when the third derivative dominates over the fourth, we have stable oscillations and vice versa. The expression for $ \qq $ has a Newton-like structure, where the inverse Hessian is applied to a gradient. However, this gradient acts only on the cubic term in the Taylor expansion of the loss, not on the full objective. Note that $ \nabla_{\vv}\D^{3} \FullLoss(\params^*)[\vv]^3$ equals $ 3\D^{3}\FullLoss(\params^*)[\vv]^2 $, and thus~$ \vv_{\max} $'s scale and polarity do not affect the condition. When the condition is satisfied, step sizes slightly above $ 2/\lambda_{\max} $ produce stable periodic oscillations, whose amplitude grows with $\eta$, while smaller step sizes converge to the minimum. Conversely, if the condition is not met, any step size larger than $ 2/\lambda_{\max} $ leads the iterates to escape the small neighborhood of the minimum. A key assumption underlying our result is that the Hessian at the minimum is strictly positive definite. While this assumption may not hold in general for machine learning problems, particularly in neural network training, it is often satisfied when explicit regularization, such as $\ell_2$ regularization, is applied. A proof outline is given in Sec.~\ref{Sec:Derivation}.

\begin{figure}[t]
    \subfigure[Graphs of $ \FullLoss_{\beta = 0.1} $ (left) and $ \FullLoss_{\beta=0.5} $ (right)][b]{%
        \includegraphics[trim={0.3cm 0 1.3cm 0},clip,width=0.268\linewidth]{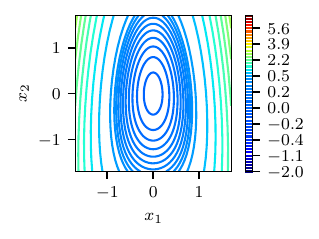}
        \includegraphics[trim={1.1cm 0 0.15cm 0},clip,width=0.292\linewidth]{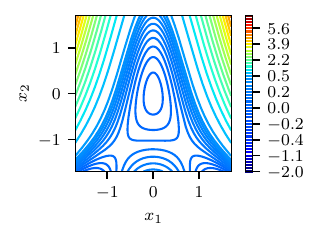}
        \label{fig:GD_func_graphs}
    }%
    \subfigure[$\params_{T}$ versus $ \beta $][b]{%
        \centering
        \includegraphics[width=0.44\linewidth]{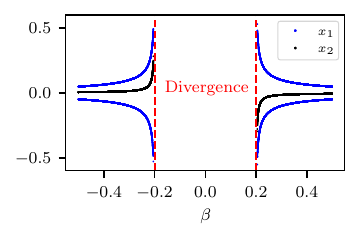}
        \label{fig:GD Oscillations}
    }
    \caption{{\bfseries Demonstration of Thm.~\protect\ref{Thm:Stable oscillations of GD}.} Consider $\FullLoss_{\beta} ( x_1, x_2) = \frac{1}{2}x_1^2 +\frac{1}{10}x_2^2 + \beta x_1^2 x_2 + \frac{1}{10} x_1^4 $, whose linear stability threshold under GD at the vicinity of the local minimizer $ \params^* = (0,0) $ is $\eta_{\lin} = 2$. According to Thm.~\protect\ref{Thm:Stable oscillations of GD}, GD at the edge of stability oscillates stably around $\params^*$ if and only if $|\beta| > 0.2$ (see App.~\protect\ref{app:GD analytic example}). Panel~\protect\subfigref{fig:GD_func_graphs} plots $\FullLoss_{\beta}$ near $\params^*$ for $\beta = 0.1$ and $ \beta = 0.5 $, highlighting the asymmetry introduced by the cubic term. Panel~\protect\subfigref{fig:GD Oscillations} shows the long-term value $\params_T$ across a range of $\beta$. When $|\beta| > 0.2$, GD converges to a stable period-$2$ cycle, whereas for $|\beta| < 0.2$ the iterates diverge. This validates that condition~\protect\eqref{eq:condition for stable oscillations} precisely captures the transition from stability to instability.
    }
    \label{fig:GD demonstration}
\end{figure}

To illustrate Thm.~\ref{Thm:Stable oscillations of GD} and its difference from~\eqref{eq:GD sufficient condition}, we consider the following example. Let
\begin{equation}
    \FullLoss_{\beta} ( x_1, x_2) = \frac{1}{2}x_1^2 +\frac{1}{10}x_2^2 + \beta x_1^2 x_2 + \frac{1}{10} x_1^4,
\end{equation}
with a minimizer at $\params^* = (0,0)$. Figure~\ref{fig:GD_func_graphs} depicts $\FullLoss _\beta$ near the minimum for $\beta = 0.1$ and $ \beta = 0.5$. The linear stability threshold of GD at $\params^*$ is $\eta_{\lin} = 2$, and the condition for stable oscillations \eqref{eq:condition for stable oscillations} simplifies to $|\beta| > 0.2$. Figure~\ref{fig:GD Oscillations} presents the accumulation points of GD iterates obtained numerically for a range of~$\beta$. When $|\beta| > 0.2$, GD's iterates converge to stable period-$2$ cycles, whereas for $|\beta| < 0.2$, the iterates diverge. This example demonstrates that Thm.~\ref{Thm:Stable oscillations of GD} captures the precise phase transition from stable to unstable oscillations. In contrast, the condition in \eqref{eq:GD sufficient condition} is not satisfied for any value of $ \beta \in \R$, leading to the fact that the hypothesis is incorrect. Full details are given in App.~\ref{app:GD analytic example}.

Although \eqref{eq:GD sufficient condition} does not correctly characterize the behavior of GD, it remains a natural condition to examine. It is therefore interesting to ask what this condition actually implies. Let $ \{ \vv_i  \}_{i = 1}^d $ denote the eigenvectors of the Hessian $ \nabla^2 \FullLoss ( \params^* ) $. Then our exact condition \eqref{eq:condition for stable oscillations} can be expressed in the following equivalent form (see App.~\ref{app:Alternative form of GD theorem}).
\begin{equation}\label{eq:alternative form of GD stability condition}
    \text{(Equivalent condition to Theorem~\ref{Thm:Stable oscillations of GD})} \quad
     {3 \sum_{i=1}^d \frac{ \big( \D^3 \FullLoss  ( \params^* ) [\vv_{\max} ]^2 [\vv_i ] \big) ^2 }{ \D^2 \FullLoss  ( \params^* ) [\vv_{i} ]^2  } > \D^4  \FullLoss ( \params^* )[\vv_{\max} ]^{4}}.
\end{equation}
This representation clarifies how the univariate condition \eqref{eq:GD univariate exact stability condition} generalizes to the multivariate setting. Here, the cubic term is decomposed into contributions from the eigendirections of the Hessian. Specifically, each term $ \D^3 \FullLoss(\params^*) [\vv_{\max}]^2[\vv_i] $ can be interpreted as the directional derivative of the top eigenvalue of the Hessian\footnote{This follows from standard eigenvalue perturbation theory and the assumption that the top eigenvalue is simple.} $ \lambda_{\max} $  along $ \vv_i $. This quantity measures how the sharpness of the loss changes under perturbations in the eigendirection $ \vv_i $. These squared derivatives are then normalized by the curvature $ \lambda_{i} = \D^2 \FullLoss  ( \params^* ) [\vv_{i} ]^2 $ in the corresponding eigendirection, and summed over all directions. Consequently, stable oscillations are determined not solely by the behavior in the sharpest direction, but by the combined contributions from all eigendirections. In the univariate setting, the sum reduces to a single term, recovering exactly the condition in~\eqref{eq:GD univariate exact stability condition}.

This form also interprets the hypothesized condition \eqref{eq:GD sufficient condition}, which accounts only for a single term ($ \vv_i = \vv_{\max} $) in the series. Since all summands are nonnegative, satisfaction of \eqref{eq:GD sufficient condition} implies \eqref{eq:condition for stable oscillations}.
\begin{corollary}[Sufficient Condition for Stable Oscillations]
    Under the assumptions of Thm.~\ref{Thm:Stable oscillations of GD}, if~\eqref{eq:GD sufficient condition} holds or $ \D^4  \FullLoss ( \params^* )[\vv_{\max} ]^{4} < 0 $ then a stable period-$2$ cycle exists in a neighborhood of $ \params^* $.
\end{corollary}
Thus, \eqref{eq:GD sufficient condition} is a sufficient condition, not necessary as prior work hypothesized. Overall, our result generalizes, corrects, clarifies, and interprets prior work, while providing a complete and exact characterization of the stability of oscillations in GD near minima in the general multivariate setting.

\section{Stability of Nonlinear Dynamics in SGD}\label{Sec:Stability of nonlinear dynamics in SGD}
In this section, we present our results on the stability of nonlinear dynamics in SGD. Let $\loss_i : \R^d \! \to \R$ be analytic for all $i\in [n]$. We define the loss function and its batch approximation as
\begin{equation}\label{eq:SGD loss}
    \FullLoss(\params)=\frac{1}{n}\sum_{i=1}^n \loss_i(\params),
    \qquad \text{and} \qquad
    \stochasticLoss_{\batch}(\params) = \frac{1}{B}\sum_{i \in \batch } \loss_i(\params),
\end{equation}
where $ \batch \subseteq [n] $ is a batch (set) of size $|\batch| = B$. The iterates of SGD are given by
\begin{equation}\label{eq:UpdateRule}
	\params_{t+1} = \params_t - \eta \nabla \stochasticLoss_{\batch_t} (\params_t).
\end{equation}
Here, $\batch_t$ refers to a stochastic batch sampled at iteration $t$. We assume that the batches $ \{ \batch_t \} $ are sampled independently across iterations, and without replacement within each batch. Thus, samples do not repeat within a batch, but may appear in multiple batches across different iterations.

Our analysis focuses on the dynamics of SGD near interpolating minimizers. This setting has been extensively studied by prior work, particularly in the context of dynamical stability and overparameterized models \citep{wu2018sgd,ma2021on,mulayoff24a}.
\begin{definition}[Interpolating Minimizer]\label{def:Interpolating minimizer}
    We say $ \params^* \in \R^d $ is an interpolating minimizer of $ \FullLoss $ if
    \begin{equation}
        \forall i \in [n] \qquad \nabla \loss_i(\params^*) = \zeroVec \qquad  \text{and}  \qquad \nabla^2 \loss_i(\params^* ) \succ \zeroVec \ \text{ (positive definite)}.
    \end{equation}
\end{definition}

To gain intuition about the stability of SGD near interpolating minimizers, it is useful to examine the dynamics of the iterates across all possible batches. Concretely, consider running GD separately on every batch. For a given step size, some batches may converge, while others may exhibit stable oscillations or even diverge. To capture the average behavior of the algorithm, we adopt the notion of stability in expectation \citep{ma2021on}. A popular instance of this approach is the mean-square stability \citep{wu2018sgd}, whose step size threshold in the linear setting aggregates curvature information from all samples \citep{mulayoff24a}. However, as shown in Sec.~\ref{Sec:Background}, nonlinear dynamics behave differently. Instability of even a single batch can be enough to cause the mean to diverge.

\begin{theorem}[Necessary Condition]\label{Thm:Unstable oscillations in SGD}
    Let $ \params^* $ be an interpolating minimizer of $ \FullLoss $, $ \params_0 \in \R^d$, and $ \batch_* $ be a batch of size $ B $. Denote GD's iterates with step size $ \eta $ over $ \stochasticLoss_{\batch_*} $ by $ \params^{\scriptscriptstyle (\batch_*)}_t $. If
    \begin{equation}\label{eq:superlinear} 
       \sqrt[{\scriptstyle t}]{\left\| \params^{\scriptscriptstyle (\batch_*)}_t - \params^* \right\|} \underset{t \to \infty}{\longrightarrow} \infty,
    \end{equation}
    then SGD's iterates $ \{ \params_t \} $ of \eqref{eq:UpdateRule} with step size $ \eta $ diverge in expectation, i.e., $ \E \big[ \| \params_t - \params^* \| \big] \underset{t \to \infty}{\longrightarrow} \infty$.
\end{theorem}
In simple terms, this theorem states that if GD on even a single batch diverges at a rate higher than linear, then SGD as a whole will also diverge in expectation (see proof in App.~\ref{Appendix:Proof of necessary condition for SGD}). Section~\ref{Sec:Warmup oscillations in gradient descent} provides a concrete example, where GD’s iterates on the function $f_+$ in \eqref{eq:normal forms of flip bifurcation} diverge superlinearly (see App.~\ref{Appendix::Worst case condition proof}). Consequently, if the finite-sum loss $\FullLoss$ contains a batch loss ${\stochasticLoss_{\batch}} = f_+$, then SGD will diverge in expectation. This is the underlying principle behind the observation in Sec.~\ref{Warmup:Stability threshold of stochastic gradient descent}.

Theorem~\ref{Thm:Unstable oscillations in SGD} has an important implication for the stability threshold of SGD. Suppose the iterates reach a neighborhood of a minimizer $\params^*$, and let $ \smash{\eta_{\batch} = {2}/{\lambda_{\max}(\nabla^2 \stochasticLoss_{\batch}(\params^*))}} $ denote the linear stability threshold of a batch loss $\smash{\stochasticLoss_{\batch}}$. Clearly, within a sufficiently small neighborhood of \(\params^*\), divergence of GD can occur only if $\eta \geq \eta_{\batch}$. Notably, if the condition for stable oscillations in Thm.~\ref{Thm:Stable oscillations of GD} is violated, then superlinear divergence may already occur at the threshold $\eta = \eta_{\batch}$. In this case, the stability threshold of SGD is effectively capped by $\eta_{\batch}$. This naturally raises the following question. Under what conditions can we guarantee SGD stability? The result below addresses this point.
\begin{theorem}[Sufficient Condition]\label{Thm:Sufficient condition for SGD}
    Let $ \params^* $ be an interpolating minimizer of $ \FullLoss $, and consider SGD's iterates~\eqref{eq:UpdateRule} denoted by $ \{ \params_t \} $. If
    \begin{equation}\label{eq:Sufficient condition}
        \eta < {\min_{ \batch : |\batch| = B } \frac{2}{\lambda_{\max} \big( \nabla^2 \stochasticLoss_{\batch} (\params^*) \big) } },
    \end{equation}
    then there exists a neighborhood $ \{ \params_0 \! : \! \| \params_0 - \params^* \| < \rho \}$ s.t. $\E\big[ \| \params_t - \params^* \|_k^k \big] \rho^{-k} \! \underset{t \to \infty}{\longrightarrow} 0  $ for all even~$ k $.
\end{theorem}
This result shows that if the step size is linearly stable with respect to all batches, then the full nonlinear dynamics of SGD are stable in expectation (proof in Sec.~\ref{sec:Sufficient condition for stability of SGD}). As demonstrated in Sec.~\ref{Warmup:Stability threshold of stochastic gradient descent}, this sufficient condition can also be necessary in certain cases. Specifically, this occurs when unstable oscillations leading to superlinear divergence arise in batches with low linear stability thresholds.

\section{Derivations}\label{Sec:Derivation}
In the following, we derive our results. Specifically, in Sec.~\ref{sec:flip bifurcation} we formulate GD dynamics at the edge of stability as a flip bifurcation to obtain Thm.~\ref{Thm:Stable oscillations of GD}. In Sec.~\ref{sec:Sufficient condition for stability of SGD}, we derive Thm.~\ref{Thm:Sufficient condition for SGD}. The proof for Thm.~\ref{Thm:Unstable oscillations in SGD} is given in App.~\ref{Appendix:Proof of necessary condition for SGD}.

\subsection{Gradient Descent Oscillations as a Flip Bifurcation}\label{sec:flip bifurcation}
In this section, we give a brief review of bifurcations and formulate GD's dynamics in this framework. For a comprehensive overview of bifurcations, see \cite{kuznetsov1998elements}. Consider the parameter-dependent nonlinear system $ \params_{t+1} = \GDtransform_{\eta}(\params_t) $ with fixed point $ \params^* $, \ie, $ \GDtransform_{\eta}(\params^*) = \params^* $. In general, a bifurcation occurs when a change in a system parameter alters the qualitative behavior of the dynamics, often by changing the stability of a fixed point. The special case of flip bifurcation, also called period-doubling, happens when the fixed point $ \params^* $ loses stability as the parameter $ \eta$ changes, and a period-$2$ cycle emerges. Mathematically, let us define the linear stability threshold $\eta_{\lin} $ on the parameter $\eta$ as the value for which the dominant\footnote{Dominant eigenvalue is an eigenvalue that has maximal absolute value. Here we assume that it is unique.} eigenvalue of the Jacobian $\D \GDtransform_{\eta_{\lin}}(\params^*)$ equals $-1$. A flip bifurcation occurs when this eigenvalue crosses $ -1 $ on the real axis as $ \eta $ passes through $ \eta_{\lin} $.

In this case, for $ \eta < \eta_{\lin} $, the fixed point $ \params^* $ is linearly stable, and if the iterates enter a sufficiently small neighborhood of $\params^*$, they will be attracted to it. In fact, when all eigenvalues of the Jacobian lie strictly inside the unit disk, the iterates converge to \(\params^*\). Yet, when $ \eta $ is slightly above~$ \eta_{\lin} $, the fixed point $ \params^* $ is no longer stable, and a period-$2$ cycle appears as
\begin{equation}
    \GDtransform_{\eta}\big(\params^{(1)}\big) = \params^{(2)}, \qquad \text{and} \qquad \GDtransform_{\eta}\big(\params^{(2)}\big) = \params^{(1)}.
\end{equation}
The stability of the resulting period-$2$ cycle is governed by the coefficient of the cubic term in the corresponding normal form of the bifurcation. This form provides a canonical (standard) dynamics to which any flip bifurcation can be reduced. Concretely, consider the dynamics along the one-dimensional critical manifold, tangent to the dominant eigenvector of the Jacobian. Then the dynamics on this manifold can be transformed into \citep[Sec.~5.4]{kuznetsov1998elements}
\begin{equation}
    {\xi_{t+1} = -\xi_{t} + C_0 \xi^3_{t} + O\big(\xi_{t}^4\big),}
\end{equation}
where $ C_0 $ is the first coefficient in this normal form. When $ C_0 > 0 $, the resulting cycle is stable (supercritical bifurcation), and the dynamics in the long run will alternate between $\params^{(1)}$ and~$\params^{(2)}$. Whereas for $ C_0 < 0 $, the cycle is unstable (subcritical bifurcation) and the iterates will diverge from~$ \params^* $. The expression for $ C_0 $, involving high-order derivatives of $ \GDtransform_{\eta} $ at $ \params^* $, is given in App.~\ref{Appendix:Condition for stable oscillations in GD}.

We now apply this framework to derive Thm.~\ref{Thm:Stable oscillations of GD}. In the context of GD's iterates near a minimum, the dynamics evolve according to the update rule
\begin{equation}\label{eq:Definition of GDtransform}
    \params_{t+1} = \params_t - \eta \nabla \FullLoss(\params_t) \triangleq \GDtransform_{\eta}(\params_t),
\end{equation}
where $ \FullLoss $ is an objective function to be minimized, and $ \eta $ is the step size. Minimizers of~$ \FullLoss $ are fixed points of~$ \GDtransform $, as the gradient vanishes at these points. The Jacobian of the GD map is
\begin{equation}
    \D \GDtransform_{\eta}(\params) = \IdentityMat - \eta \nabla^2 \FullLoss(\params).
\end{equation}
Note that the eigenvalues of the Jacobian are given by $ \{1-\eta \lambda_{i}(\nabla^2 \FullLoss ) \} $. Let $ \params^* $ be a minimizer of~$ \FullLoss $, then the corresponding linear stability threshold is the well-known linear stability threshold
\begin{equation}
    \eta_{\lin} = \frac{2}{\lambda_{\max}\big(\nabla^2 \FullLoss(\params^*)\big)}.
\end{equation}
Thus, as $ \eta$ exceeds $ \eta_{\lin} $, the dominant eigenvalue of the Jacobian crosses $-1$ on the real axis, matching the scenario of the flip bifurcation. Assuming $ \nabla^2 \FullLoss(\params^*) $ is positive definite and $ \lambda_{\max}(\nabla^2 \FullLoss(\params^*)) $ has multiplicity one, the stability of the bifurcating period-$2$ cycle is governed by~$ C_0 $. Overall, we see that oscillations near a minimum $ \params^* $ are stable if and only if $ C_0 $ is positive. In App.~\ref{Appendix:Condition for stable oscillations in GD} we show that a positive coefficient is equivalent to the condition in Thm.~\ref{Thm:Stable oscillations of GD}.

\subsection{Sufficient Condition for Stability of SGD}\label{sec:Sufficient condition for stability of SGD}
In this section, we derive Thm.~\ref{Thm:Sufficient condition for SGD}. SGD update rule with step size $ \eta $ is given by
\begin{equation}\label{eq:update_rule2}
    \params_{t+1} = \params_t - \eta \nabla \stochasticLoss_{\batch_t}(\params_t) \triangleq \SGDtransform_{\batch_t}(\params_t),
\end{equation}
where $ \SGDtransform_{\batch_t}: \R^d \to \R^d $ is the SGD map. As $ \params^* $ is an interpolating minimizer of $ \FullLoss $, we have
\begin{equation}
    \SGDtransform_{\batch_t}(\params^*) = \params^* \ \text{ w.p. } \ 1.
\end{equation}
Since $ \{ \loss_i \} $ are analytic, we can use the Taylor expansions of $ \SGDtransform_{\batch_t} $ at $ \params^* $ to get
\begin{equation}
    \SGDtransform_{\batch_t}(\params) = \SGDtransform_{\batch_t}(\params^*) + \sum_{k=1}^{\infty} \frac{1}{k!} \mD^k \SGDtransform_{\batch_t}(\params^*)  (\params-\params^*)^{\otimes k},
\end{equation}
where $ \mD^k $ is the $ k $th order derivative in \emph{matrix form} (not to be confused with $\D^k$). Let
\begin{equation}
    \Delta\params^k_t \triangleq (\params_t-\params^*)^{\otimes k} \in \R^{d^k}
\end{equation}
be the $k$th Kronecker power of the distance to the minimum\footnote{The first power $ \Delta\params_t^1$ is denoted simply $ \Delta\params_t $.}. Then from the update rule in \eqref{eq:update_rule2}
\begin{equation}\label{eq:dynamical_system2}
    \E\left[\Delta\params_{t+1}\right] = \E\left[\sum_{k=1}^{\infty} \frac{1}{k!} \mD^k \SGDtransform_{\batch_t}(\params^*) \Delta\params^k_t \right] = \sum_{k=1}^{\infty} \frac{1}{k!} \E \left[ \mD^k \SGDtransform_{\batch_t}(\params^*)  \right] \E \left[ \Delta\params^k_t\right],
\end{equation}
where we used that $\batch_t$ is sampled independently of the history up to time $t$. Thus, $ \{ \mD^k \SGDtransform_{\batch_t}(\params^*) \}_{k=1}^{\infty} $ is independent of $ \params_t$.
We see that the evolution of the first moment of the distance to the minimum, $ \E[\Delta\params] $, depends \emph{linearly} on all higher-order moments $ \{ \E[\Delta\params^k] \}_{k=1}^{\infty} $. Consequently, analyzing the stability of SGD in expectation requires studying the joint dynamics of all moments. In App.~\ref{Appendix:Computation of the operator blocks}, we show that the evolution of the~$k$th moment over time is
\begin{equation}\label{eq:evolution of the moments}
    \E \left[ \Delta\params^k_{t+1} \right]
    = \E \left[ \left(\Delta\params_{t+1} \right)^{\otimes k} \right]
    = \E \left[ \left(\sum_{p=1}^{\infty} \frac{1}{p!} \mD^p \SGDtransform_{\batch_t}(\params^*) \Delta\params^p_t\right)^{\otimes k} \right]
    = \sum_{p=k}^{\infty} \bPsi_{k,p} \E \left[ \Delta\params^p_t \right],
\end{equation}
where an explicit expression for $ \bPsi_{k,p} \in \R^{ d^k \times d^p} $ is given in App.~\ref{Appendix:Computation of the operator blocks}. Once again, we obtain a linear relation between the moments at successive times. This motivates us to express the mapping from $ \{ \E[\Delta\params^k_{t}] \}_{k=1}^\infty $ to $ \{ \E[\Delta\params^k_{t+1}] \}_{k=1}^\infty $ as a linear operator on the infinite-dimensional Hilbert space $ \ell_2 $. This formulation is meaningful only if the sequence has finite norm and the operator is bounded. To ensure this, we introduce a radius $ \rho > 0 $ and analyze a scaled version of the moments. Let
\begin{equation}\label{eq:normalized moments}
    \bar{\bmu}^k_t
    \triangleq \E \left[ \left(\frac{\params_t-\params^*}{\rho}\right)^{\otimes k} \right]
    = \rho^{-k} \E \left[ \Delta \params^k_{t} \right].
\end{equation}
Therefore,
\begin{equation}
    \bar{\bmu}_{t+1}^k
    = \rho^{-k} \E \left[ \Delta \params^k_{t+1} \right]
    = {\sum_{p=k}^{\infty} \rho^{-k} \bPsi_{k,p} \E \left[ \Delta\params^p_t \right]
    = \sum_{p=k}^{\infty} \rho^{p-k} \bPsi_{k,p} \E \left[ \rho^{-p} \Delta \params^p_t \right]
    = \sum_{p=k}^{\infty} \rho^{p-k} \bPsi_{k,p} \bar{\bmu}_{t}^p}.
\end{equation}
Define the linear operator $ \bPsi_{\rho} $ acting on the Hilbert space $ \ell_2 $ and the vector of moments $ \bar{\bmu}_t $ as
\begin{equation}
    \bar{\bmu}_{t}
    = \begin{bmatrix}
        \bar{\bmu}_{t}^1\\ \bar{\bmu}_{t}^2\\ \bar{\bmu}_{t}^3\\ \vdots
    \end{bmatrix}
    \qquad \text{and}
    \qquad
    \bPsi_{\rho} = \begin{bmatrix}
        \bPsi_{1,1} & \rho \bPsi_{1,2}  & \rho^2\bPsi_{1,3}  & \cdots \\
        \zeroVec & \bPsi_{2,2} & \rho \bPsi_{2,3} &  \cdots \\
        \zeroVec & \zeroVec & \bPsi_{3,3} &  \cdots \\
        \vdots & \vdots & \vdots & \ddots & 
    \end{bmatrix},
\end{equation}
then
\begin{equation}\label{eq:basic_linear_equation}
    \bar{\bmu}_{t+1} = \bPsi_{\rho} \bar{\bmu}_{t}.
\end{equation}
This relation is valid only when $ \bPsi_{\rho} $ is bounded. Intuitively, taking smaller values of $ \rho $ suppresses the off-diagonal blocks through the factors $ \rho^{p-k} \ (p>k)$, which helps ensure that the operator is bounded. Assuming the operator is bounded, \eqref{eq:basic_linear_equation} unfolds as $ \bar{\bmu}_{t} = \bPsi_{\rho}^t \bar{\bmu}_{0} $. To determine the neighborhood in which the analysis applies, observe that $ \bar{\bmu}_{0} \in \ell_2 $, and thus must be square-summable.
\begin{equation}
    {\| \bar{\bmu}_{0} \|^2
    = \sum_{k=1}^{\infty} \left\| \bar{\bmu}^k_{0} \right\|^2
    = \sum_{k=1}^{\infty} \left\|\left(\frac{\params_0 - \params^*}{\rho}\right)^{\otimes k} \right\|^2
    = \sum_{k=1}^{\infty}\left( \frac{\|\params_0-\params^* \|}{\rho} \right)^{2 k}}.
\end{equation}
The above expression is finite if and only if $ \| \params_{0}-\params^* \| < \rho $, which defines the neighborhood around the minimum where our analysis applies. We see that choosing a smaller $\rho$ to ensure boundedness of the operator correspondingly shrinks this neighborhood around the minimizer.

For stable dynamics in expectation, the linear system in \eqref{eq:basic_linear_equation} must be stable. Note that under $ \bPsi_{\rho} $, moment vectors map naturally to moment vectors. Thus, to characterize the exact stability threshold, we would need $ \bPsi_{\rho} $ to be contractive on this restricted set. Instead, we relax this constraint to obtain a sufficient condition, requiring stability for any vector in $ \ell_2 $. In App.~\ref{Appendix:OperatorBounding}, we prove that under the condition \eqref{eq:Sufficient condition}, there exists a value of $ \rho > 0 $ ensuring boundedness of the operator. Then, in App.~\ref{Appendix:SpectralAnalysis} we show that once the operator is bounded, its spectral radius is strictly less than one. Therefore, $ \| \bar{\bmu}_{t} \| \to 0 $ as $ t $ tends to infinity (see App.~\ref{Appendix:Spectral analysis introduction}). Hence, each normalized moment also tends to zero elementwise, \ie, $ \bar{\bmu}^k_{t} \to \zeroVec $. Since $ \bar{\bmu}^k_{t} $ contains all degree-$k$ monomials of the components of~$ \Delta\params_t $, denoted $ \{ \Delta\params_{t,i} \}_{i=1}^d $, summing over the subset of single-variable terms
\begin{equation}
    \sum_{i=1}^d \E \left[ \left(\Delta\params_{t,i} \right)^k \right] \rho^{-k}
    = \E \left[ \sum_{i=1}^d  \left(\params_{t,i}-\params^{*}_{i} \right) ^k \right]\rho^{-k}
    \underset{t \to \infty }{\longrightarrow} 0 .
\end{equation}
Restricting this to even order moments (even $ k $), we get
$ \E \big[ \|\params_t-\params^* \|^{k}_{k} \big]\rho^{-k} \underset{t \to \infty }{\longrightarrow} 0 $.

\section{Related Work}\label{Sec:Related Work}

\paragraph{Bifurcation, Oscillations and EoS in GD.}
\citet{cohen2021gradient} examined the behavior of GD on neural networks, and found that it typically occurs at the edge of stability (EoS). \citet{wang2022analyzing} proved progressive sharpening for a two-layer network and analyzed the EoS dynamics through four phases, depending on the change in the sharpness value. \citet{zhu2022understanding} gave a simple example that exhibits EoS. \citet{ma2022beyond} analyzed the EoS under the assumption that the loss has subquadratic growth. They begin with the general univariate case and extend their analysis to higher dimensions under a specific structural assumption. Within this framework, GD with a fixed step size exhibits only two possible outcomes: it either enters a periodic cycle or converges to a minimum; no other behavior exists. Namely, only a supercritical bifurcation is possible in their setting. This contrasts with our framework, in which we derive conditions that determine the type of bifurcation: supercritical or subcritical. Moreover, the subquadratic-growth assumption does not capture many practical machine learning scenarios, \emph{e.g.}, regression with the square loss. \citet{ahn2022understanding} introduced tools such as directional smoothness and relative progress ratio to formalize loss fluctuations. Using these metrics, the authors showed that the stability of GD arises from structured oscillations. Although they provide explicit formulas for these quantities, they do not derive conditions under which these oscillations are stable. \cite{damian2023selfstabilization} showed how GD self-stabilizes. Specifically, they demonstrated that during the momentary divergence of the iterates along the sharpest eigenvector direction of the Hessian, the iterates also move along the negative direction of the gradient of the sharpness, effectively taking a gradient step with respect to $\lambda_{\max}(\nabla^2 \FullLoss )$, which can explain the stabilization of the sharpness around $ 2/\eta $. Yet, they do not provide conditions for the descent of the sharpness. \citet{kreisler2023gradient} and \citet{song2023trajectory} proved that under EoS, different GD trajectories align on a specific bifurcation diagram independent of initialization. \cite{chen2023beyond} studied the same setting we consider and derived the exact condition for stable oscillations for univariate optimization. However, as we discuss in Sec.~\ref{Sec:Oscillations in gradient descent}, it was unclear to them how this condition extends to the multivariate setting. \citet{chen2024stability} examined GD dynamics on quadratic loss from stability up to the chaos phase. \citet{ghosh2025learning} analyzed the dynamics of deep linear networks, focusing on $2$-period cycles, while showing that oscillations occur within a small subspace, where the dimension of the subspace is controlled by the step size.

\paragraph{Stability of SGD.}
Empirically, \citet{keskar2016large,jastrzkebski2017three,jastrzebski2018on,Jastrzebski2020The} showed that SGD with a large step size or small batch size leads to flatter minima. \citet{cohen2021gradient} found that with large batches, the sharpness behaves similarly to full-batch gradient descent. \citet{gilmer2022a} studied how the curvature of the loss affects the training dynamics in multiple settings. Theoretically, \citet{wu2018sgd} analyzed stability in the mean-square sense and provided an implicit sufficient condition. \citet{granziol2022learning} used random matrix theory to characterize the maximal stable learning rate as a function of batch size, under certain assumptions on Hessian noise. \citet{velikanov2023a} studied SGD with momentum and derived an implicit upper bound on the learning rate using spectrally expressible approximations and a moment-generating function. \citet{ma2021on} investigated higher-order moments of SGD and established an implicit necessary and sufficient stability condition. \citet{wu2022alignment} proposed a necessary condition based on an alignment property, though a general analytic bound for this property is missing. \citet{ziyin2023probabilistic} examined stability in probability rather than in mean square, showing that SGD can in theory converge with high probability to linearly unstable minima for GD, \ie, where $ \eta \gg 2/\lambda_{\max}( \nabla^2 \FullLoss) $. However, this prediction was not observed empirically. \citet{mulayoffNeurips} considered non-differentiable minima and gave a necessary condition for strong stability, meaning SGD remains within a ball around the minimum. Finally, \citet{mulayoff24a} derived the exact stability criterion in a closed-form expression for the linearized dynamics.

Additionally, \citet{pmlr-v139-liu21ad} analyzed the covariance matrix of the stationary distribution of iterates near minima, and \citet{ziyin2022strength} extended these results by deriving an implicit relation between this covariance and that of the gradient noise. However, both works leave open the question of when the dynamics actually converge to a stationary state. Recently, \citet{lee2023a} examined the stability of SGD along its trajectory and established an explicit exact condition for objective decrease via a descent lemma in expectation.

\section{Conclusion, Limitations, and Future Directions}\label{Sec:Conclusion}
In this paper, we examined nonlinear effects in the stability of GD and SGD. For GD, we derived an explicit condition characterizing the stability of oscillations at the edge of stability, while clarifying and correcting prior work. For SGD, we showed that the instability of even a single batch can be sufficient to render the entire dynamics unstable in expectation, implying that stability is dictated by the worst-case batch rather than by an average effect. Finally, we proved that if the step size is stable with respect to all batches, then all moments of the full nonlinear SGD dynamics remain stable in a neighborhood of the minimizer. Together, these results reveal that nonlinear effects can fundamentally reshape the stability landscape compared with standard linear analyses.

\paragraph{Limitations and future directions.} Our analysis of oscillations in GD focuses on isolated minima, where the Jacobian of the dynamical system has a single critical eigenvalue. In deep learning, however, minima often form low-dimensional manifolds with multiple near-critical directions. This can lead to richer local dynamics, including combinations of fold and flip behaviors, and a more complex stability picture. Extending our results to this setting, 
potentially via generalized fold-flip bifurcations \citep{kuznetsov2004fold}, is an important direction for future work.

For SGD, our analysis is restricted to interpolating minima. This choice stems from the difficulty of defining stability of the full dynamics with respect to non-interpolating minima. In such cases, SGD cannot converge exactly to the minimizer; even if the dynamics are stable, the algorithm exhibits an inherent bias in expectation \citep{defossez2015averaged}. As a result, it is unclear whether the final distance from the minimizer reflects this bias or whether the algorithm converges to a different minimum altogether. In this work, we adopt the convergence-in-expectation perspective, but we believe that developing a more principled definition of stability for the non-interpolating case remains an important direction for future research.


\bibliography{StableMinimaRef}

\clearpage

\appendix
{\noindent\LARGE\bfseries Appendix}
\renewcommand{\thesection}{\Roman{section}} 

\section{Notations}\label{app:notations and Kronecker}
Throughout the paper and our derivations, we use the following notations.
{\renewcommand{\arraystretch}{1}
\begin{table}[h]
	\begin{tabular}{c l}
		$ a $ & Lower case non-bold letters for scalars \\ 
		$ \boldsymbol{a} $ & Lower case bold letters for vectors \\ 
		$ \boldsymbol{A} $ 					& Upper case bold for matrices \\
		$ \boldsymbol{A}^{\transpose} $ 	& Transpose of $ \boldsymbol{A} $ \\
		$ \otimes $ 						& Kronecker product \\
        $ ^{ \otimes k } $                    & $k$'th Kronecker power \\
        $  f \circ g$                    & Composition of $ f $ and $ g $, \emph{i.e.}, $ f(g(\params)) $ \\
		$ \E $ 								& Expectation \\
        $ \Prob $ 								& Probability function \\
		$ \|{\boldsymbol{a}}\| $ 			& Euclidean norm of $\boldsymbol{a} $ \\
		$ \|{\boldsymbol{A}}\| $ 			& Operator norm of $\boldsymbol{A} $ (top singular value in case of a matrix) \\
		$ r(\boldsymbol{A}) $ 			& Spectral radius of $\boldsymbol{A} $ \\
        $ \sigma(\boldsymbol{A}) $ 			& Spectrum of $\boldsymbol{A}$ (singular value in case of a matrix) \\
        $ \lambda(\boldsymbol{A}) $ 			& Eigenvalue of $\boldsymbol{A} $ \\
        $ \FullLoss $ 							& Objective function to minimize (empirical risk) \\
		$ \loss_{i} $ 							& Loss function of the $i$th sample \\
		$ \params $ 						& Parameters vector of the objective function \\
		$ \params^{*} $ 					& Minimum point of the objective function \\
		$ d $ 								& Dimension of $ \params $ \\
		$ n $ 								& Number of training samples  \\
		$ \eta $ 							& Step size \\
		$ B $ 								& Batch size \\
		$ \rho $ 							& Analysis (convergence) radius for SGD \\
        $\IdentityMat_d $                      & The $d \times d $ identity matrix (when the dimensions are clear, the subscript is omitted) \\
        $\D^k $ & The $k $th order derivative in multilinear form, see \citet[App.~A.6.7]{ben2021lectures} \\
        $ \mD^k $ & the $ k $th order derivative in matrix form \\
        $\GDtransform$ & The gradient descent operator defined in \eqref{eq:Definition of GDtransform}\\
        $\SGDtransform_{\batch}$ & The stochastic gradient descent operator on batch $ \batch $ defined in \eqref{eq:update_rule2}\\
        $ \bar{\bmu}^k_t $ & The normalized $k$th moment of SGD dynamic at time $t$, defined in \eqref{eq:normalized moments}
	\end{tabular}
	\caption{Table of notations}
\end{table}
}

Additionally, we use the following properties of the Kronecker product during our derivations. For any matrices $\MM_1,\MM_2,\MM_3,\MM_4$,
\begin{align*}
    \big( \MM_1 \otimes \MM_2 \big)^\transpose  & =  \MM_1^\transpose \otimes \MM_2^\transpose \tag{P1} \label{eq:KroneckerProperty2}, \\
    \big( \MM_1 \otimes \MM_2 \big) \big( \MM_3 \otimes \MM_4 \big) & = \big( \MM_1 \MM_3 \big) \otimes \big( \MM_2 \MM_4 \big) \tag{P2}. \label{eq:KroneckerProperty3}
\end{align*}

\section{Background on Linear Stability of SGD}\label{app:Background on linear stability}
Analyzing the full dynamics of SGD can be hard. Therefore, many works opt to study the linearized dynamics near minima \citep{wu2018sgd,ma2021on,mulayoffNeurips,mulayoff24a}, as it is common in the analysis of nonlinear systems. In our paper, we focus on interpolating minimizers, defined in Def.~\ref{def:Interpolating minimizer}. In this case, the linearized dynamics is defined below.
\begin{definition}[Linearized dynamics]\label{def:Linearization}
    Let $ \FullLoss $ from \eqref{eq:SGD loss}, and 
    $ \params^* $ be its interpolating minimizer, s.t. $ \FullLoss $ is twice differentiable  at $\params^* $. Then the linearized dynamics of SGD near $ \params^*  $ are given by
    \begin{equation}\label{eq:linearized dynamics}
        \tilde{\params}_{t+1} = \tilde{\params}_{t} - \frac{\eta}{B} {\sum_{i \in  \batch_t }} \nabla^2 \loss_i(\params^* ) (\tilde{\params}_t - \params^* ).
    \end{equation}
\end{definition}
The linearized dynamics can be viewed as SGD on the second-order approximation of $ \FullLoss $ at~$ \params^* $,
\begin{equation}\label{eq:TaylorLtilde}
    \tilde{\FullLoss}(\params) = \FullLoss(\params^*) + {\frac{1}{2}} (\params-\params^* )^\transpose \nabla^2 \FullLoss (\params^*) (\params - \params^*).
\end{equation}
Therefore, linear dynamics analysis is exact only when $ \{ \loss_i \} $ are all quadratic potentials.

There are a few traditional ways to define the convergence of random processes, such as the iterates of SGD. One prominent choice is to use the mean square sense of convergence to define stability. For univariate optimization, the mean square linear stability threshold is as follows. Generalization to higher dimensions and non-interpolating minima can be found in \citet{mulayoff24a}.
\begin{theorem}[Univariate Linear Stability Threshold, {\citet{wu2018sgd}}]\label{thm:linear dynamics univariate}
Let $ \loss_i: \R \to \R $ be twice differentiable functions and let $ \param^* $ be an interpolating minimum of the loss, i.e.,
\begin{equation}
    \forall \ 1\leq i \leq n \qquad \loss_i'(\param^*) = 0 \qquad \text{and} \qquad h_i \triangleq  f_i''(\param^*) >0.
\end{equation}
Define
\begin{equation}
    h = \frac{1}{n}\sum_{i = 1}^{n} h_i, \qquad s^2 = \frac{1}{n}\sum_{i = 1}^{n} (h_i -h)^2, \qquad \text{and} \qquad p = \frac{n-B}{B(n-1)}.
\end{equation}
Consider the iterates of the linearized SGD $ \{ \tilde{\param}_t \}$ in \eqref{eq:linearized dynamics}. Then, $ \E[(\tilde{\param}_t-\param^*)^2] $ is bounded if and only if $ \eta \leq \eta_{\lin}$, where
\begin{equation}\label{eq:SGD linear threshold}
    \eta_{\lin} \triangleq \frac{2h}{h^2+ps^2}.
\end{equation}
\end{theorem}
From this result, we see that the linear stability threshold $ \eta_{\lin} $ takes into account the sharpness of all functions $ \{ \loss_i \} $.
When the batch size $ B $ equals one, we get $ p = 1 $ and then
\begin{equation}
    \eta_{\lin} = \frac{2h}{h^2 + s^2} = 2\frac{ \sum_{i = 1}^{n} h_i}{\sum_{i = 1}^{n} h^2_i}.
\end{equation}

\section{Proof of Proposition~\protect\ref{prop:Worst case condition}}\label{Appendix::Worst case condition proof}
Let 
\begin{equation}
    \loss_+(\param) = \frac{1}{2}\param^2 + \frac{1}{4}\param^4.
\end{equation}
Here we show that when GD is applied on $ \loss_+ $ with step size $ \eta > 2 $, its iterates diverge at a rate higher than linear. In this case, Thm.~\ref{Thm:Unstable oscillations in SGD} tells us that SGD dynamics also diverge.

The GD map on $ \loss_+ $ is
\begin{equation}
    \psi(\param) = \param - \eta \loss_+'(\param) = (1-\eta)\param -\eta \param^3.
\end{equation}
Define
\begin{equation}
    \psi^t = \underbrace{\psi \circ \cdots \circ \psi \circ \psi}_{t \text{ times}}.
\end{equation}
Assume that $\eta >2$, and note that for all $\param \in \R$
\begin{equation}
    |\psi(\param)| = |(1-\eta)\param -\eta \param^3| = (\eta-1)|\param|+\eta |\param|^3 = |\psi(|\param|)|.
\end{equation}
Let $ \tilde{\psi}(\param) = |\psi|(|\param|) $, then
\begin{equation}
    |\psi^t (\param_0)| = \tilde{\psi}^t(|\param_0|).
\end{equation}
Since $ \tilde{\psi}: \R_+ \to \R_+  $ is monotonically increasing on $\R_+$, we have that its composition $ \tilde{\psi}^t $ of any order is also monotonically increasing. Furthermore, we can bound $\tilde{\psi}$ from below with
\begin{equation}
    \tilde{\psi}(\param) = (\eta-1)|\param|+\eta |\param|^3 \geq \max\{(\eta-1)|\param|,\eta |\param|^3\} \triangleq \varphi(\param).
\end{equation}
Thus,
\begin{equation}
   | \param_t | = |\psi^t (\param_0)| \geq \varphi^t (\param_0) \triangleq \chi_t.
\end{equation}
Again, $ \varphi : \R_+ \to \R_+ $ is monotonically increasing, and therefore any composition with itself is also monotonically increasing on $\R_+ $. Obviously,
\begin{equation}
    \varphi (\param) \geq (\eta-1)|\param| \qquad \text{and} \qquad \varphi (\param) \geq \eta |\param|^3.
\end{equation}
Then we can bound $ \chi_t $ by
\begin{equation}
    \chi_t \geq (\eta-1)^t |\param_0|.
\end{equation}
Since $ \eta >2 $, there exists $ T \in \N $ such that
\begin{equation}
    \chi_T \geq (\eta-1)^T|\param_0|>2.
\end{equation}
Now, for all $ t > T$ we have
\begin{equation}
    \chi_{t} = \varphi^{t-T} (\chi_T).
\end{equation}
Here, we will use the second bound, \ie, $ \varphi (\param) \geq \eta |\param|^3 $, $ t-T $ times as
\begin{align}
    \chi_{t}
    & = \underbrace{\eta | \eta | \cdots  \eta |\chi_T|^3 \cdots |^3|^3 }_{t-T \text{ times}} \nonumber \\
    & = \eta^{(3^{t-T}-1)/2}|\chi_T|^{3^{t-T}}\nonumber \\
    & \geq 2^{(3^{t-T}-1)/2}2^{3^{t-T}} \nonumber \\
    & = 2^{(3^{t-T+1}-1)/2}.
\end{align}
Therefore, $ \chi_t $ diverges with a superlinear rate, and so does $ | \param_t | $.

\section{Condition for Stable Oscillations in GD}\label{Appendix:Condition for stable oscillations in GD}
In Sec.~\ref{sec:flip bifurcation} we formulate the oscillations of GD as a flip bifurcation. We saw that the first coefficient $ C_0 $ controls the stability of the oscillations. For a general nonlinear map $ \GDtransform_{\eta}(\params) $, this coefficient is given by \citep[Sec.~5.4]{kuznetsov1998elements}
\begin{equation}
    C_0 = \frac{1}{6}\left\langle \uu ,  \D^3 \GDtransform_{\eta}(\params^*)[\vv, \vv , \vv ]  \right\rangle
    - \frac{1}{2} \left\langle \uu , \D^2 \GDtransform_{\eta}(\params^*)  [\vv, \pp ] \right\rangle,
\end{equation}
where $ \uu $ and $ \vv $ are normalized left and right eigenvectors of the Jacobian $ \D \GDtransform_{\eta}(\params^*) $ corresponding to the eigenvalue $ -1 $, such that $ \langle \uu ,\vv \rangle  = 1$, and
\begin{equation}
    \pp = \big[ \D \GDtransform_{\eta}(\params^*) - \IdentityMat \big]^{-1} \D^2 \GDtransform_{\eta}(\params^*)[\vv, \vv ].
\end{equation}
We would like to write these expressions in terms of the loss function $ \FullLoss $ for the GD dynamics. In this case, $ \GDtransform_{\eta}(\params) = \params - \eta \nabla \FullLoss(\params) $, and we have that the Jacobian is
\begin{equation}
    \D \GDtransform_{\eta}(\params^*) = \IdentityMat - \eta \nabla^2 \FullLoss(\params^*).
\end{equation}
Since this Jacobian is symmetric, $ \vv$ equals $\uu $. Moreover, it is easy to see that $ \vv $ is the top eigenvector of the Hessian, corresponding to $ \lambda_{\max}(\nabla^2 \FullLoss(\params^*) ) $. Additionally,
\begin{equation}
    \D^2 \GDtransform_{\eta}(\params^*)[\vv, \vv ] = - \eta \D^3 \FullLoss (\params^* ) [\vv, \vv ] = - \frac{\eta}{3} \nabla_{\vv} \D^3 \FullLoss (\params^* ) [\vv, \vv, \vv] .
\end{equation}
Thus,
\begin{align}
    \pp
    & = \big[ - \eta \nabla^2 \FullLoss(\params^*) \big]^{-1}  \left( - \frac{\eta}{3} \nabla_{\vv} \D^3 \FullLoss (\params^* ) [\vv, \vv, \vv]  \right) \nonumber \\
    & = \frac{1}{3}\big[ \nabla^2 \FullLoss(\params^*) \big]^{-1} \nabla_{\vv} \D^3 \FullLoss (\params^* ) [\vv, \vv, \vv] \nonumber \\
    & = \frac{1}{3} \qq,
\end{align}
where
\begin{equation}
    \qq = \big[ \nabla^2 \FullLoss(\params^*) \big]^{-1} \nabla_{\vv} \D^3 \FullLoss (\params^* ) [\vv, \vv, \vv].
\end{equation}
Next, we have
\begin{equation}
    \left\langle \uu , \D^2 \GDtransform_{\eta}(\params^*)  [\vv, \pp ] \right\rangle = \frac{1}{3}\left\langle \vv , \D^2 \GDtransform_{\eta}(\params^*)  [\vv, \qq ] \right\rangle = -\frac{\eta}{3} \D^3 \FullLoss (\params^* ) [\vv, \vv, \qq ].
\end{equation}
And the first term in $ C_0 $ is
\begin{equation}
    \left\langle \uu ,  \D^3 \GDtransform_{\eta}(\params^*)[\vv, \vv , \vv ]  \right\rangle = \left\langle \vv ,  -\eta \D^4 \FullLoss(\params^*)[\vv, \vv , \vv ]  \right\rangle = -\eta \D^4 \FullLoss(\params^*)[\vv, \vv , \vv, \vv ].
\end{equation}
Overall,
\begin{equation}
    C_0 = -\frac{\eta}{6} \D^4 \FullLoss(\params^*)[\vv, \vv , \vv, \vv ] + \frac{\eta}{6} \D^3 \FullLoss (\params^* ) [\vv, \vv, \qq ].
\end{equation}
A period-$2$ cycle near $ \params^* $ is stable if and only if $ C_0 > 0 $ \citep{kuznetsov1998elements}, which results in the condition
\begin{equation}
    \D^3 \FullLoss (\params^* ) [\vv, \vv, \qq ] > \D^4 \FullLoss(\params^*)[\vv, \vv , \vv, \vv ].
\end{equation}
Originally, the scale of $ \vv $ was important for the magnitude of $ C_0 $. However, the scale of $ \vv $ has no effect on the sign of $ C_0 $, and thus has no impact on this condition.

\section{Alternative Form of Theorem~\protect\ref{Thm:Stable oscillations of GD}}\label{app:Alternative form of GD theorem}
Our condition for stable oscillations in Theorem~\ref{Thm:Stable oscillations of GD} is
\begin{equation}
    \D^3 \FullLoss( \params^* ) \big[ \vv_{\max} \big]^2 \big[ \qq \big] > \D^4 \FullLoss( \params^* ) \big[ \vv_{\max} \big]^{4},
\end{equation}
where
\begin{equation}
    \qq \triangleq \Big[ \nabla^2 \FullLoss( \params^* )  \Big]^{-1} \nabla_{\vv} \D^{3} \FullLoss( \params^* ) \big[ \vv \big]^3  \Big|_{\vv = \vv_{\max}}.
\end{equation}
Note that 
\begin{equation}
    \nabla_{\vv}  \D^{3} \FullLoss( \params^* ) \big[ \vv \big]^3  \Big|_{\vv = \vv_{\max}} =  3 \D^{3} \FullLoss( \params^* ) [\vv_{\max}]^2. 
\end{equation}
Let $ \{ ( \lambda_i, \vv_i ) \}_{i = 1}^{d} $ denote the eigenpairs of the Hessian at the minimum $ \params^* $, then
\begin{equation}
    \Big[ \nabla^2 \FullLoss( \params^*  )  \Big]^{-1} =  \sum_{i = 1}^d \frac{1}{\lambda_{i} } \vv_{i}  \vv_{i}^{\transpose}.
\end{equation}
Thus,
\begin{equation}
    \qq 
    = 3 \sum_{i = 1}^d \frac{1}{\lambda_{i} } \vv_{i}  \vv_{i}^{\transpose}\D^{3} \FullLoss( \params^* ) [\vv_{\max}]^2 
    = 3 \sum_{i = 1}^d \frac{\D^{3} \FullLoss( \params^* ) [\vv_{\max}]^2 [\vv_{i}] }{\lambda_{i} } \vv_{i} .
\end{equation}
Therefore,
\begin{align}
    \D^{3} \FullLoss( \params^* ) \big[ \vv_{\max} \big]^2 \big[ \qq \big]
    & = 
    \D^{3} \FullLoss( \params^* )\big[ \vv_{\max} \big]^2 \left[
    3 \sum_{i = 1}^d \frac{\D^{3} \FullLoss( \params^* ) [\vv_{\max}]^2 [\vv_{i}] }{\lambda_{i} } \vv_{i} \right] \nonumber \\
    & = 3 \sum_{i = 1}^d  \frac{ \left( \D^{3} \FullLoss( \params^* ) [\vv_{\max}]^2 [\vv_{i}] \right)^2 }  {\lambda_{i} } ,
\end{align}
where the last step is by linearity. Using the identity
\begin{equation}
    \lambda_i = \D^2 \FullLoss  ( \params^* ) [\vv_{i} ]^2 ,
\end{equation}
we get that Theorem~\ref{Thm:Stable oscillations of GD} is equivalent to
\begin{equation}
     {3 \sum_{i=1}^d \frac{ \big( \D^3 \FullLoss  ( \params^* ) [\vv_{\max} ]^2 [\vv_i ] \big) ^2 }{ \D^2 \FullLoss  ( \params^* ) [\vv_{i} ]^2  }> \D^4  \FullLoss ( \params^* )[\vv_{\max} ]^{4}}.
\end{equation}

\section{Analytic Example of Theorem~\protect\ref{Thm:Stable oscillations of GD}}\label{app:GD analytic example}
Let us consider the dynamics of GD on the two-dimensional function
\begin{equation}
    \FullLoss ( \param_1, \param_2) = \frac{1}{2}x_1^2 +\frac{1}{10}x_2^2 + \beta x_1^2 x_2   + \frac{1}{10} x_1^4 ,
\end{equation}
in the vicinity of the minimum $ \params^{*} = (x_1^{*},x_2^{*})^{\transpose}  = (0,0)^{\transpose} $. Here, the Hessian's eigenvalues at the minimum $ \params^* $ are $  \lambda_{\max} = 1, \ \lambda_{\min} = 0.2 $ and the corresponding eigenvectors are 
\begin{equation}
    \vv_{\max} = (1,0)^{\transpose} , \qquad  \vv_{\min} = (0,1)^{\transpose}.
\end{equation}
Let us use the equivalent form of Thm.~\ref{Thm:Stable oscillations of GD}, given in \eqref{eq:alternative form of GD stability condition}.
The first term in the series on the left-hand side is
\begin{equation}
    3\frac{ \big( \D^3 \FullLoss  ( \params^* ) [\vv_{\max} ]^2 [\vv_{\max}] \big) ^2 }{ \D^2 \FullLoss  (\params^*) [\vv_{\max} ]^2  }
    = 3 \frac{ (  \partial_{ \vv_{ \max } }^3  \FullLoss (0,0) ) ^2 }{ \lambda_{\max} }  =
    3 \frac{ (  \partial_{ x_{1} }^3  \FullLoss (0,0) )^2 }{ \lambda_{\max} }  = 0.
\end{equation}
The second term is the series is
\begin{equation}
    3\frac{ \big( \D^3 \FullLoss  ( \params^* ) [\vv_{\max} ]^2 [\vv_{\min}] \big) ^2 }{ \D^2 \FullLoss  (\params^*) [\vv_{\min} ]^2  }
    = 3 \frac{ ( \partial_{ \vv_{\min} } \partial_{ \vv_{\max} }^2  \FullLoss (0,0) ) ^2 }{ \lambda_{\min} }  =
    3 \frac{ ( \partial_{ x_{2} } \partial_{ x_{1} }^2  \FullLoss (0,0) ) ^2 }{ \lambda_{\min} }  = 
    3 \frac{ \left( 2 \beta \right)^2} { 0.2 } = 60 \beta^2.
\end{equation}
The right-hand side of the condition is
\begin{equation}
    \D^4  \FullLoss ( \params^* )[\vv_{\max} ]^{4} = 
     \partial_{ \vv_{\max} }^4  \FullLoss (0,0) =  \partial_{ {x_1} }^4  \FullLoss (0,0) = \frac{4!}{10} = 2.4.
\end{equation}
First, we see that the condition proposed by \citet{chen2023beyond},
\begin{equation}
    0 = 3 \frac{ \big(  \D^3 \FullLoss  ( \params^* ) [\vv_{\max} ]^3 \big) ^2 }{ \D^2 \FullLoss  ( \params^* ) [\vv_{\max} ]^2  }> \D^4  \FullLoss ( \params^* )[\vv_{\max} ]^{4} =  2.4,
\end{equation}
does not hold for any value of $ \beta $. However, our exact condition for stable oscillations, 
\begin{equation}
     {3 \sum_{i=1}^d \frac{ \big( \D^3 \FullLoss  ( \params^* ) [\vv_{\max} ]^2 [\vv_i ] \big) ^2 }{ \D^2 \FullLoss  ( \params^* ) [\vv_{i} ]^2  } > \D^4  \FullLoss ( \params^* )[\vv_{\max} ]^{4}}
\end{equation}
is $ 60 \beta^2 > 2.4 $, which reduces to $  | \beta | > 0.2 $. Then, our theory asserts that whenever $  | \beta | > 0.2 $, the oscillations are stable; otherwise, the oscillations are unstable.

\section{Proof of Theorem~\protect\ref{Thm:Unstable oscillations in SGD}}\label{Appendix:Proof of necessary condition for SGD}
Let $ \{ \batch_{i} \}_{i = 1}^N $ be all possible different batches of size $ B $ from the dataset $ \{ \loss_i \}_{i =1}^n $, where $ N = \binom{n}{B} $. Recall that $ \SGDtransform_{\batch} $ denotes the of GD transform on batch $ \batch $, \ie, taking a single gradient step with respect to $ \stochasticLoss_\batch $ (see~\eqref{eq:update_rule2}). Moreover, let $\SGDtransform^t_{\batch} $ denote the application of $ \SGDtransform_{\batch} $ for $ t $ times. Namely,
\begin{equation}
    \SGDtransform^t_{\batch} = \underbrace{\SGDtransform_{\batch} \circ \cdots \circ \SGDtransform_{\batch}  \circ \SGDtransform_{\batch} }_{t \text{ times}} .
\end{equation}
For a stochastic batch $ \batch_t $, $ \SGDtransform_{\batch_t} $ is distributed uniformly over $ \{ \SGDtransform_{\batch_i } \} $, \ie, for any $ \params \in \R^d $
\begin{equation}
    \SGDtransform_{\batch_i }(\params) \sim \mathcal{U} \left( \left\{ \SGDtransform_{\batch_i }(\params) \right\}_{i = 1}^N \right).
\end{equation}
Given an initial point $ \params_0 \in \R^d $, assume that for some batch $ \batch_{i^*} $ (with index $ i^* $) GD's iterates, denoted by $ \{ \params^{\scriptscriptstyle (\batch_{i^*})}_t \} $, diverge with superlinear rate. That is
\begin{equation}\label{eq:super linear appendix}
    \big\| \params^{\scriptscriptstyle (\batch_{i^*})}_t - \params^* \big\|^{\frac{1}{t}} \underset{t \to \infty}{\longrightarrow} \infty,
\end{equation}
Using our notation, we have $ \params^{\scriptscriptstyle (\batch_{i^*})}_t = \SGDtransform^t_{\batch_{i^*} }(\params_0)$. Let us look at the expectation of the distance between SGD iterates $ \{ \params_{t} \}  $ from \eqref{eq:UpdateRule} and the minimizer $ \params^* $ 
\begin{align}
    \E\left[ \| \params_{t}-\params^* \|  \right]
    & = \frac{1}{N^t} \sum_{(i_1,i_2,\ldots, i_t) \in \{ 1,\ldots,N \}^t} \big\| \SGDtransform_{\batch_{i_t} } \circ \cdots \circ \SGDtransform_{\batch_{i_2} } \circ \SGDtransform_{\batch_{i_1} } (\params_0) - \params^* \big\|  \nonumber \\
    & \geq  \frac{1}{N^t} \big\| \SGDtransform_{\batch_{i_t} } \circ \cdots \circ \SGDtransform_{\batch_{i_2} } \circ \SGDtransform_{\batch_{i_1} } (\params_0) - \params^* \big\|  \Big|_{i_1 = i_2 = \cdots = i_t = i^*} \nonumber \\
    & = \frac{1}{N^t} \big\| \SGDtransform^t_{\batch_{i^*} } (\params_0) - \params^* \big\| \nonumber \\
    & = \exp \left\{  \log\left(\big\| \SGDtransform^t_{\batch_{i^*} } (\params_0) - \params^* \big\| \right) - t \log(N)  \right\} \nonumber \\
    & = \exp \left\{ t \left[ \frac{1}{t} \log\left(\big\| \SGDtransform^t_{\batch_{i^*} } (\params_0) - \params^* \big\| \right) -  \log(N)  \right] \right\} \nonumber \\
    & = \exp \left\{ t \left[ \log\left(\big\| \params^{\scriptscriptstyle (\batch_{i^*})}_t - \params^* \big\|^{\frac{1}{t}} \right) -  \log(N)  \right] \right\} \underset{t \to \infty}{\longrightarrow} \infty.
\end{align}

\section{Introduction to Spectral Analysis of Linear Operators}\label{Appendix:Spectral analysis introduction}
Let us start with the following definitions.
\begin{definition}[Operator Norm]
    Let $ \A $ be a linear operator over a vector space $ V $, then its operator norm is given by
    \begin{equation}
        \| \A \| = \inf \{ c\geq 0 \ : \ \| \A \vv \| \leq c \| \vv \| \quad \text{for all} \ \vv \in V \}.
    \end{equation}
\end{definition}
\begin{definition}[Spectrum]
    Let $ \A $ be a linear operator over a Banach space $ V $, then its spectrum is given by
    \begin{equation}
        \sigma(\A) = \{ \lambda \in \C \ : \A -\lambda\IdentityMat \ \text{is not bijective} \},
    \end{equation}
    where $ \IdentityMat $ is the identity operator.
\end{definition}
\begin{definition}[Spectral Radius]
    The spectral radius of an operator $ \A $ is given by
    \begin{equation}
        r(\A) = \sup_{\lambda \in \sigma(\A)} |\lambda|.
    \end{equation}
\end{definition}
Consider the following linear system
\begin{equation}
    \bmu_{t+1} = \A \bmu_{t},
\end{equation}
where $ \A $ is a bounded linear operator. We want a condition such that the iterates are either bounded or converging. Here, we can unfold the equation to get an explicit formula for any $ \bmu_t $ as
\begin{equation}
    \bmu_t = \A^t \bmu_0.
\end{equation}
A naive way to ensure convergence, \ie, $ \bmu_t \to \zeroVec $ as $ t \to \infty $, is by taking the operator norm of $ \A $ to be less than one, \ie, $ \| \A \| < 1 $. Then
\begin{equation}
    \left\| \bmu_t \right\| = \left\| \A^t \bmu_0 \right\| \leq  \left\| \A \right\|^t \left\| \bmu_0 \right\| \to 0 .
\end{equation}
However, this is quite restrictive and will give us a loss condition. Note that we only like to know if $ \| \A^t \| $ is bounded or shrinks to zero. In special cases, we can easily compute~$ \A^t $. For example, in the finite-dimensional case, where $ \A = \PP \DD \PP^{-1} $ is diagonalizable (\eg, symmetric or normal). Then,
\begin{equation}
    \bmu_t = \A^t \bmu_0 = \left( \PP \DD \PP^{-1}  \right)^t \bmu_0 = \PP \DD^t \PP^{-1} \bmu_0.
\end{equation}
Here, the system is stable if and only if the spectral radius of $ \A $ is less or equal to one. If it is strictly less than one, then $ \bmu_t \to \zeroVec $. 

In the general case, we can use Gelfand's formula for bounded linear operators on Banach spaces. Let $ r(\A) $ denote the spectral radius of $ \A $, then Gelfand's formula is
\begin{equation}
    r(\A) = \lim_{t \to \infty} \left\| \A^t \right\|^{\frac{1}{t}} = \inf_{t \in \N} \left\| \A^t \right\|^{\frac{1}{t}}.
\end{equation}
From this formula, we can see that if $ r(\A)<1 $, then $ \bmu_t \to \zeroVec $.

\section{Computation of the Operator Blocks}\label{Appendix:Computation of the operator blocks}
In this section, we give the missing steps from \eqref{eq:evolution of the moments}. To this end, denote
\begin{equation}\label{eq:derivative in matrix form}
    \YY_{t,k} = \frac{1}{k!} \mD^k \SGDtransform_{\batch_t}(\params^*)  \in \R^{d \times d^k} .
\end{equation}
Then, the evolution over time of the $ k $th moment is
{\allowdisplaybreaks
\begin{align}
    & \E \left[ \left(\sum_{p=1}^{\infty} \frac{1}{p!} \mD^p \SGDtransform_{\batch_t}(\params^*) \Delta\params^p_t\right)^{\otimes k} \right] \nonumber \\
    & = \E\left[ \left(\sum_{p=1}^{\infty} \YY_{t,p}  \Delta\params_t^{ p} \right)^{\otimes k} \right] \nonumber \\
    & = \E\left[ \sum_{p=k}^{\infty} \; \sum_{ \substack{ 1\leq \kappa_1, \cdots, \kappa_{k} \leq p-k+1 \\ \kappa_1+\kappa_2+\cdots+\kappa_k = p }} \left( \YY_{t,\kappa_1} \Delta\params_t^{ \kappa_1} \right) \otimes \left( \YY_{t,\kappa_2} \Delta\params_t^{ \kappa_2} \right) \otimes \cdots \otimes \left( \YY_{t,\kappa_k} \Delta\params_t^{ \kappa_k} \right)  \right] \nonumber \\
    & = \E\left[ \sum_{p=k}^{\infty} \; \sum_{ \substack{ 1\leq \kappa_1, \cdots, \kappa_{k} \leq p-k+1 \\ \kappa_1+\kappa_2+\cdots+\kappa_k = p }} \left( \YY_{t,\kappa_1} \otimes  \cdots \otimes \YY_{t,\kappa_k} \right) \left( \Delta\params_t^{ \kappa_1} \otimes\cdots \otimes\Delta\params_t^{ \kappa_k} \right)  \right] \nonumber \\
    & = \E\left[ \sum_{p=k}^{\infty} \; \sum_{ \substack{ 1\leq \kappa_1, \cdots, \kappa_{k} \leq p-k+1 \\ \kappa_1+\kappa_2+\cdots+\kappa_k = p }} \left( \YY_{t,\kappa_1} \otimes \YY_{t,\kappa_2}  \otimes \cdots \otimes \YY_{t,\kappa_k} \right) \left( \Delta\params_t^{ \sum_{i=1}^{k} \kappa_i} \right)  \right] \nonumber \\
    & = \E\left[ \sum_{p=k}^{\infty} \; \sum_{ \substack{ 1\leq \kappa_1, \cdots, \kappa_{k} \leq p-k+1 \\ \kappa_1+\kappa_2+\cdots+\kappa_k = p }} \left( \YY_{t,\kappa_1} \otimes \YY_{t,\kappa_2}  \otimes \cdots \otimes \YY_{t,\kappa_k} \right)  \Delta\params_t^{ p }  \right] \nonumber \\
    & = \E\left[ \sum_{p=k}^{\infty} \;  \left( \sum_{ \substack{ 1\leq \kappa_1, \cdots, \kappa_{k} \leq p-k+1 \\ \kappa_1+\kappa_2+\cdots+\kappa_k = p }} \YY_{t,\kappa_1} \otimes \YY_{t,\kappa_2}  \otimes \cdots \otimes \YY_{t,\kappa_k} \right)  \Delta\params_t^{ p }  \right] \nonumber \\
    & = \sum_{p=k}^{\infty} \; \E\left[ \sum_{ \substack{ 1\leq \kappa_1, \cdots, \kappa_{k} \leq p-k+1 \\ \kappa_1+\kappa_2+\cdots+\kappa_k = p }} \YY_{t,\kappa_1} \otimes \YY_{t,\kappa_2}  \otimes \cdots \otimes \YY_{t,\kappa_k} \right] \E\left[   \Delta\params_t^{ p }  \right] \nonumber \\
    & = \sum_{p=k}^{\infty} \bPsi_{k,p} \E \left[  \Delta\params^p_t \right],
\end{align}}
where 
\begin{equation}\label{eq:operator block definition}
    \bPsi_{k,p} = \E\left[ \sum_{ \substack{ 1\leq \kappa_1, \cdots, \kappa_{k} \leq p-k+1 \\ \kappa_1+\kappa_2+\cdots+\kappa_k = p }} \YY_{t,\kappa_1} \otimes \YY_{t,\kappa_2}  \otimes \cdots \otimes \YY_{t,\kappa_k} \right] \in \R^{d^k \times d^p},
\end{equation}

\section{Bounding the Operator}\label{Appendix:OperatorBounding}
In this section, we assume that the condition of Thm.~\ref{Thm:Sufficient condition for SGD} holds. Then, we show that there exists a $ \rho > 0  $ such that $ \bPsi_{\rho} $ is bounded. For this, we use the following result (see proof in App.~\ref{Appendix:BoundnessSufficientCondition}).
\begin{theorem}\label{thm:boundness test}
    Let $ \TT $ be an operator defined on $ \ell_2 $ space. Denote by $ \{ \TT_{i,j} \}$ a division of $\TT $ into blocks, such that $ \forall i,j \; \TT_{i,j} \in \R^{d_i \times d_j}$ where $ \{ d_i \}_{i=1}^\infty $ is a some sequence. Assume that 
    \begin{equation}
        \forall j \in \N \qquad \sum_{i = 1}^\infty \|\TT_{i,j}\| \leq \alpha \qquad \text{and} \qquad \forall i \in \N \qquad \sum_{j = 1}^\infty \|\TT_{i,j}\| \leq \beta.
    \end{equation}
    Then $ \TT $ is a bounded linear operator and
    \begin{equation}
        \| \TT \| \leq \sqrt{\alpha \beta}.
    \end{equation}
\end{theorem}
Let us apply Thm.~\ref{thm:boundness test} to $ \bPsi_{\rho} $. Using the definition of the blocks $ \{ \bPsi_{k,p} \} $ given in \eqref{eq:operator block definition} we have
\begin{align}
    \| \bPsi_{k,p} \|
    & = \left\| \E\left[ \sum_{ \substack{ 1\leq \kappa_1, \cdots, \kappa_{k} \leq p-k+1 \\ \kappa_1+\kappa_2+\cdots+\kappa_k = p }} \YY_{t,\kappa_1} \otimes \YY_{t,\kappa_2}  \otimes \cdots \otimes \YY_{t,\kappa_k} \right] \right\| \nonumber \\
    & \leq \E\left[\sum_{ \substack{ 1\leq \kappa_1, \cdots, \kappa_{k} \leq p-k+1 \\ \kappa_1+\kappa_2+\cdots+\kappa_k = p }} \big\| \YY_{t,\kappa_1} \otimes \YY_{t,\kappa_2}  \otimes \cdots \otimes \YY_{t,\kappa_k} \big\| \right] \nonumber \\
    & = \E\left[\sum_{ \substack{ 1\leq \kappa_1, \cdots, \kappa_{k} \leq p-k+1 \\ \kappa_1+\kappa_2+\cdots+\kappa_k = p }} \big\| \YY_{t,\kappa_1} \big\| \big\| \YY_{t,\kappa_2}  \big\| \cdots  \big\| \YY_{t,\kappa_k} \big\| \right],
\end{align}
where $ \YY_{t, p } $ is given in \eqref{eq:derivative in matrix form}. Let $ \{ \batch_{m} \}_{m = 1}^N $ be all possible different batches of size $ B $ from the dataset $ \{ \loss_m \}_{m =1}^n $, where $ N = \binom{n}{B} $. Since $ \{ \loss_m \} $ are analytic and $ \{ \stochasticLoss_{\batch_{m}} \} $ are finite sum losses, then also $  \{ \SGDtransform_{\batch_{m}} \}  $ are analytic. Then, using Gevrey class theory, for each batch $ \batch_{m} $ there exists $ C_{m} > 0 $ such that
\begin{equation}
    \max_{i,j} \left| \Big[ \mD^p \SGDtransform_{\batch_m}(\params^*) \Big]_{i,j} \right| \leq C_m^{p+1} p! \qquad  \forall p \geq 1 ,
\end{equation}
where $ [ \mD^p \SGDtransform_{\batch_m}(\params^*)]_{i,j} $ are the elements of the matrix $ \mD^p \SGDtransform_{\batch_m}(\params^*) $, which are all the (mixed) partial derivatives of degree $ p $. Setting
\begin{equation}
    C = \max_{m \in [N]} C_m,
\end{equation}
we get for a random batch $ \batch_{t} $
\begin{equation}
    \max_{i,j} \left| \Big[ \mD^p \SGDtransform_{\batch_t}(\params^*) \Big]_{i,j} \right| \leq C^{p+1} p! \quad \text{w.p.} \ \ 1 \qquad  \forall p \geq 1 .
\end{equation}
Now that we have a uniform bound on all elements in the matrix, we can bound its norm. Specifically, it is well known that for a matrix $ \A \in \R^{ m \times n } $ with elements $ | \A_{i,j} | \leq M $, we have $ \| \A \| \leq M \sqrt{m n}  $ (a simple application of Thm.~\ref{thm:boundness test} can give this result as well). Using this, and the fact that $ \mD^p \SGDtransform_{\batch_t}(\params^*)  \in \R^{d \times d^p} $ we get
\begin{equation}
    \|\YY_{t,p}\| = \frac{1}{p!} \left\| \mD^p \SGDtransform_{\batch_t}(\params^*) \right\| \leq C^{p+1} d^{\frac{p+1}{2}}  \quad \text{w.p.} \ \ 1.
\end{equation}
Define
\begin{equation}
    Q_{t,p} = \begin{cases}
        \|\YY_{t,1}\| , & p = 1, \\[5pt]
        C^{p+1} d^{\frac{p+1}{2}}, & \text{otherwise}.
    \end{cases}
\end{equation}
Then for all $ p \geq 1 $ and $ t \in \N $
\begin{equation}
    \|\YY_{t,p}\| \leq Q_{t,p} \qquad \text{w.p. 1},
\end{equation}
and
\begin{equation}
    \|\bPsi_{k,p} \|
    \leq \E\left[\sum_{ \substack{ 1\leq \kappa_1, \cdots, \kappa_{k} \leq p-k+1 \\ \kappa_1+\kappa_2+\cdots+\kappa_k = p }}  Q_{t,\kappa_1} Q_{t,\kappa_2} \cdots Q_{t,\kappa_k} \right] .
\end{equation}
Let us apply Thm.~\ref{thm:boundness test} on $ \bPsi_{\rho} $, while assuming that $ \rho < \frac{1}{C \sqrt{d}} $. For the sum of the row block we have
\begin{align} \label{eq:row sum of A tilde2} 
    \sum_{p = k}^{\infty} \rho^{p-k}  \|\bPsi_{k,p}\|
    & \leq \sum_{p = k}^{\infty} \rho^{p-k} \E\left[\sum_{ \substack{ 1\leq \kappa_1, \cdots, \kappa_{k} \leq p-k+1 \\ \kappa_1+\kappa_2+\cdots+\kappa_k = p }}  Q_{t,\kappa_1} Q_{t,\kappa_2} \cdots Q_{t,\kappa_k} \right] \nonumber \\
    & = \rho^{-k} \E\left[ \sum_{p = k}^{\infty} \left( \sum_{ \substack{ 1\leq \kappa_1, \cdots, \kappa_{k} \leq p-k+1 \\ \kappa_1+\kappa_2+\cdots+\kappa_k = p }}  Q_{t,\kappa_1} Q_{t,\kappa_2} \cdots Q_{t,\kappa_k} \right) \rho^{p} \right] \nonumber \\
    & = \rho^{-k} \E \left[ \left( \sum_{p =1}^{\infty} Q_{t,p} \rho^{p} \right)^k  \right] \nonumber \\
    & = \rho^{-k} \E \left[ \left( \|\YY_{t,1}\|\rho + \sum_{p = 2 }^{\infty}C^{p+1} d^{\frac{p+1}{2}} \rho^{p} \right)^k  \right] \nonumber \\
    & = \rho^{-k} \E \left[ \left( \|\YY_{t,1}\|\rho +C \sqrt{d} \sum_{p = 2 }^{\infty} \left( C \sqrt{d} \rho\right)^{p}\right)^k  \right] \nonumber \\
    & = \rho^{-k} \E \left[ \left( \|\YY_{t,1}\|\rho +C \sqrt{d} \frac{C^2 d \rho^2 }{1-C \sqrt{d} \rho}\right)^k \right] \nonumber \\
    & = \E \left[ \left( \|\YY_{t,1}\| +\frac{C^3 d^{3/2} \rho }{1-C \sqrt{d} \rho} \right)^k \right],
\end{align}
where in the sixth step we used
\begin{equation}
    \sum_{p=2}^{\infty}  q^p = \frac{q^2}{1-q},
\end{equation}
for $0 < q = C \sqrt{d} \rho < 1 $. Here we assume the condition in \eqref{eq:Sufficient condition} holds and that $ \{ \nabla^2 \stochasticLoss_{\batch_i} (\params^*) \} $ have full rank. This means that for every batch $ \batch_i $
\begin{equation}
    0 <\eta \lambda_{\min} \big( \nabla^2 \stochasticLoss_{\batch_i} (\params^*) \big) < \eta \lambda_{\max} \big( \nabla^2 \stochasticLoss_{\batch_i} (\params^*) \big) < 2 .
\end{equation}
Recall that $ \mD \SGDtransform_{\batch_i}(\params^*) = \IdentityMat-\eta \nabla^2 \stochasticLoss_{\batch_i} (\params^*)$. Then it is easy to show that there exists $  \varepsilon \in (0,1) $ such that
\begin{align}
    \max_{i \in [N] } \left\|\mD \SGDtransform_{\batch_i}(\params^*) \right\|
    & = \max_{i \in [N] } \left\| \IdentityMat-\eta \nabla^2 \stochasticLoss_{\batch_i} (\params^*) \right\| \nonumber \\
    & = \max_{i \in [N] } \left\{ \max \left\{ 1 - \eta \lambda_{\min} \big( \nabla^2 \stochasticLoss_{\batch_i} (\params^*) \big) , \eta \lambda_{\max} \big( \nabla^2 \stochasticLoss_{\batch_i} (\params^*) \big) -1 \right\} \right\} \nonumber \\
    & = 1 - \varepsilon.
\end{align}
This means that 
\begin{equation}\label{eq:singular value bound}
    \|\YY_{t,1}\| \leq \max_{i \in [N] } \left\|\mD \SGDtransform_{\batch_i}(\params^*) \right\| = 1 - \varepsilon \quad \text{ w.p. } \ 1 .
\end{equation}
Note that $\frac{C^3 d^{3/2} \rho }{1-C \sqrt{d} \rho} \geq 0  $, then we can further bound \eqref{eq:row sum of A tilde2} by
\begin{equation}
    \E \left[ \left( \|\YY_{t,1}\| + \frac{C^3 d^{3/2} \rho }{1-C \sqrt{d} \rho} \right)^k \right]
    \leq \left( 1-\varepsilon +\frac{C^3 d^{3/2} \rho }{1-C \sqrt{d} \rho} \right)^k.
\end{equation}
In order for this to be bounded for any $ k \in \N $, we will require
\begin{align}
    & \qquad 1-\varepsilon +\frac{C^3 d^{3/2} \rho }{1-C \sqrt{d} \rho} < 1 \nonumber \\
    & \Leftrightarrow \qquad \frac{C^3 d^{3/2} \rho }{1-C \sqrt{d} \rho}  < \varepsilon \nonumber \\
    & \Leftrightarrow  \qquad C^3 d^{3/2} \rho + \varepsilon C \sqrt{d} \rho  < \varepsilon \nonumber \\
    & \Leftrightarrow  \qquad \rho  < \frac{\varepsilon}{C^3 d^{3/2}+ \varepsilon C \sqrt{d} }  \triangleq \rho^*.
\end{align}
Therefore, under the condition of $ \rho < \rho^* $ there exists $ \gamma \in (0,1) $ (for example, $ \gamma =  1-\varepsilon + \frac{C^3 d^{3/2} \rho }{1-C \sqrt{d} \rho}$) such that
\begin{equation}
    \sum_{p = k}^{\infty} \rho^{p-k} \|\bPsi_{k,p}\| \leq \gamma^k.
\end{equation}
This means that the rows' sum is uniformly bounded. Now, for the column sums, under the same assumptions, we get
\begin{align}\label{eq:bound block sum}
    \sup_{p} \sum_{k = 1}^{p} \rho^{p-k}\|\bPsi_{k,p}\|
    & \leq \sum_{p =1}^{\infty}\sum_{k = 1}^{p} \rho^{p-k}\|\bPsi_{k,p}\| \nonumber \\
    & =\sum_{k = 1}^{\infty} \sum_{p =k}^{\infty} \rho^{p-k}\|\bPsi_{k,p}\| \nonumber \\
    & \leq \sum_{k =1}^{\infty} \gamma^k \nonumber \\
    & = \frac{\gamma}{1-\gamma},
\end{align}
where the change of summation order in the second step is justified since all elements are non-negative. Hence, under the same assumptions, the absolute column sum is also uniformly bounded.

Overall, we see that the conditions of Thm.~\ref{Thm:Sufficient condition for SGD} are sufficient to find a neighborhood around the minimum, $ \{ \params_0 : \| \params_0 - \params^* \| < \rho \} $, such that the operator $ \bPsi_{\rho} $ is bounded. \emph{For completeness}, in App.~\ref{Appendix:BoundnessNecessaryCondition}, we show that the condition in \eqref{eq:Sufficient condition} is also necessary. Namely, if this condition is violated, $ \bPsi_{\rho} $ is not bounded.  

\section{Necessary Condition for Boundness}\label{Appendix:BoundnessNecessaryCondition}
In this section, we show that the condition in \eqref{eq:Sufficient condition} is also necessary to bound the operator $ \bPsi_{\rho} $. We bring this only to give a complete theoretical understanding, \emph{yet we do not use this derivation to prove our results.}

For $ \bPsi_{\rho} $ to be bounded, all of its submatrices must be bounded. Note that the diagonal blocks of this operator $ \{ \bPsi_{k,k} \}_{k=1}^\infty $ are independent of $ \rho $. Therefore, we should have a condition, independent of $ \rho $, for these submatrices to be bounded. These blocks are given by
\begin{equation}
    \bPsi_{k,k}
    = \E\left[ \left( \mD \SGDtransform_{\batch_t}(\params^*) \right)^{\otimes k} \right].
\end{equation}
Note that
\begin{equation}
    \mD\SGDtransform_{\batch_t}(\params^*) = \mD\left(\params -\eta  \nabla\stochasticLoss_{\batch_t}(\params)\right)\Big|_{\params = \params^*} = \IdentityMat-\eta \nabla^2\stochasticLoss_{\batch_t}(\params^*).
\end{equation}
For ease of reading and better interpretability, let
\begin{equation}
    \HH_{\batch} \triangleq \nabla^2 \stochasticLoss_{\batch}(\params^*)
\end{equation}
denote the Hessian of the batch $\batch $. Then
\begin{equation}
    \bPsi_{k,k} = \E\left[ \left( \IdentityMat - \eta \HH_{\batch_t} \right)^{\otimes k} \right].
\end{equation}
Moreover, denote by $ \HH_{\max} $ the batch that has the largest maximal eigenvalue, that is
\begin{equation}
    \HH_{\max} = \argmax_{\batch : |\batch| = B} \left\{ \lambda_{\max}\big(\HH_{\batch}\big) \right\},
\end{equation}
and by $ \vv_{\max} $ its corresponding eigenvector (normalized). Note that $ \bPsi_{k,k} $ is symmetric for all $ k \in \N $, therefore
\begin{equation}
    \left\| \bPsi_{k,k} \right\| = \max_{\uu \in \R^{d^k} \, : \, \| \uu \| = 1} \left| \uu^{\transpose} \bPsi_{k,k} \uu \right|.
\end{equation}
Since $ \| \vv_{\max}^{\otimes k} \| = \| \vv_{\max}\|^{ k}  = 1$, we have that
\begin{align}\label{eq:A_k,k norm lower bound}
    \left\| \bPsi_{k,k} \right\| 
    & \geq  \left| \left(\vv_{\max}^{\otimes k}\right)^{\transpose} \bPsi_{k,k} \vv_{\max}^{\otimes k} \right| \nonumber \\
    & = \left| \left(\vv_{\max}^{\otimes k}\right)^{\transpose} \E\left[ \left( \IdentityMat - \eta \HH_{\batch_t} \right)^{\otimes k} \right] \vv_{\max}^{\otimes k} \right| \nonumber \\
    & = \left| \E\left[\left(\vv_{\max}^{\otimes k}\right)^{\transpose} \left( \IdentityMat - \eta \HH_{\batch_t} \right)^{\otimes k}  \vv_{\max}^{\otimes k} \right] \right| \nonumber \\
    & = \left| \E\left[ \left( 1 - \eta \vv_{\max}^{\transpose}\HH_{\batch_t} \vv_{\max} \right)^{ k}  \right] \right|.
\end{align}
Assume that
\begin{equation}\label{eq:false condition}
    \eta > \frac{2}{\lambda_{\max}(\HH_{\max})}.
\end{equation}
Since $ \lambda_{\max}(\HH_{\max}) = \vv_{\max}^{\transpose}\HH_{\max} \vv_{\max} $, under the assumption above, we have that
\begin{equation}
    \Prob\left( \eta \vv_{\max}^{\transpose}\HH_{\batch_t} \vv_{\max}> 2 \right) > 0.
\end{equation}
Therefore, continuing from \eqref{eq:A_k,k norm lower bound}
\begin{align}
    \left\| \bPsi_{k,k} \right\| 
    & \geq \left| \E\left[ \left( 1 - \eta \vv_{\max}^{\transpose}\HH_{\batch_t} \vv_{\max} \right)^{ k}  \right] \right| \nonumber \\
    & = \left| \Prob\left( \eta \vv_{\max}^{\transpose}\HH_{\batch_t} \vv_{\max}> 2 \right) \E\left[ \left( 1 - \eta \vv_{\max}^{\transpose}\HH_{\batch_t} \vv_{\max} \right)^{ k} \middle| \eta \vv_{\max}^{\transpose}\HH_{\batch_t} \vv_{\max}> 2 \right]  \right. \nonumber \\
    & \quad  + \left. \Prob\left( \eta \vv_{\max}^{\transpose}\HH_{\batch_t} \vv_{\max} \leq  2 \right) \E\left[ \left( 1 - \eta \vv_{\max}^{\transpose}\HH_{\batch_t} \vv_{\max} \right)^{ k} \middle| \eta \vv_{\max}^{\transpose}\HH_{\batch_t} \vv_{\max} \leq 2 \right]   \right| \nonumber \\
    & \geq \Prob\left( \eta \vv_{\max}^{\transpose}\HH_{\batch_t} \vv_{\max}> 2 \right) \E\left[ \left(\eta \vv_{\max}^{\transpose}\HH_{\batch_t} \vv_{\max} -1 \right)^{ k} \middle| \eta \vv_{\max}^{\transpose}\HH_{\batch_t} \vv_{\max}> 2 \right]   \nonumber \\
    & \quad  - \Prob\left( \eta \vv_{\max}^{\transpose}\HH_{\batch_t} \vv_{\max} \leq  2 \right) \left|\E\left[ \left( 1 - \eta \vv_{\max}^{\transpose}\HH_{\batch_t} \vv_{\max} \right)^{ k} \middle| \eta \vv_{\max}^{\transpose}\HH_{\batch_t} \vv_{\max} \leq 2 \right]   \right|,
\end{align}
where in the second step we used the law of total expectation, and in the last step we used the triangle inequality. Since $ \params^* $ is an interpolating minimum, then $ \HH_{\batch_t} $ is PSD w.p.\ one, and $ 0 \leq \vv_{\max}^{\transpose}\HH_{\batch_t} \vv_{\max} $. Thus,
\begin{equation}
    \left|\E\left[ \left( 1 - \eta \vv_{\max}^{\transpose}\HH_{\batch_t} \vv_{\max} \right)^{ k} \middle| \eta \vv_{\max}^{\transpose}\HH_{\batch_t} \vv_{\max} \leq 2 \right]   \right| \leq 1.
\end{equation}
However,
\begin{equation}
    \E\left[ \left(\eta \vv_{\max}^{\transpose}\HH_{\batch_t} \vv_{\max} -1 \right)^{ k} \middle| \eta \vv_{\max}^{\transpose}\HH_{\batch_t} \vv_{\max}> 2 \right] \underset{k \to \infty}{\longrightarrow} \infty.
\end{equation}
This means that under the condition in \eqref{eq:false condition}, we have that $ \{ \bPsi_{k,k}  \} $ are unbounded. Therefore, a necessary condition for boundness is
\begin{equation}
    \eta \leq \frac{2}{\lambda_{\max}(\HH_{\max})}.
\end{equation}

\section{Spectral Analysis}\label{Appendix:SpectralAnalysis}
In App.~\ref{Appendix:OperatorBounding} we proved that, under the condition in \eqref{eq:Sufficient condition} of Thm.~\ref{Thm:Sufficient condition for SGD}, we can find a neighborhood $ \| \params_0-\params^* \| < \rho $ such that the operator $ \bPsi_{\rho} $ is bounded. In this section, we show that under the same condition, the spectral radius of $ \bPsi_{\rho} $, denoted by $ r(\bPsi_{\rho}) $, is less than one. To do this, we first show that the operator is compact, which means that all the non-zero elements in its spectrum are eigenvalues (point spectrum). For this end, we define the following sequence of finite rank approximations (truncations) $ \{ \bPsi^k_{\rho} \} $, comprised of the first $ k \times k $ blocks of $ \bPsi_{\rho} $. Namely,
\begin{equation}
    \bPsi^k_{\rho} = \begin{bmatrix}
        \bPsi_{1,1} & \rho \bPsi_{1,2}  & \rho^2\bPsi_{1,3} & \cdots  & \rho^{k-1} \bPsi_{1,k} & \zeroVec & \cdots \\
        \zeroVec & \bPsi_{2,2} & \rho \bPsi_{2,3} & \cdots & \rho^{k-2} \bPsi_{2,k} & \zeroVec & \cdots \\
        \zeroVec & \zeroVec & \bPsi_{3,3} & \cdots  & \rho^{k-3} \bPsi_{3,k} & \zeroVec & \cdots \\
        \vdots & \vdots & \vdots & \ddots & \vdots & \zeroVec & \cdots \\
        \zeroVec & \zeroVec & \zeroVec & \zeroVec & \bPsi_{k,k}  & \zeroVec & \cdots \\
        \zeroVec & \zeroVec & \zeroVec & \zeroVec & \zeroVec & \zeroVec & \cdots \\
        \vdots & \vdots & \vdots & \vdots & \vdots & \vdots & \ddots
    \end{bmatrix}.
\end{equation}
Furthermore, define $ \tilde{\bPsi}_{i,j} $ as the embedding of the block $ \bPsi_{i,j} $ to the full space, \ie
\begin{equation}
    \tilde{\bPsi}_{i,j} = \begin{bmatrix}
        \zeroVec  & \cdots  & \zeroVec   & \zeroVec & \zeroVec & \cdots \\
        \vdots & \ddots & \vdots  & \vdots & \vdots & \cdots \\
        \zeroVec &  \cdots  & \zeroVec & \zeroVec  & \zeroVec & \cdots \\
        \zeroVec &  \cdots  &  \zeroVec  & \bPsi_{i,j} & \zeroVec &  \cdots \\
        \zeroVec & \cdots & \zeroVec  & \zeroVec & \zeroVec & \cdots \\
        \vdots & \vdots & \vdots & \vdots & \vdots & \ddots
    \end{bmatrix},
\end{equation}
such that
\begin{equation}
    \bPsi_{\rho} = \sum_{i=1}^{\infty} \sum_{j=i}^{\infty} \rho^{j-i}\tilde{\bPsi}_{i,j} \qquad \text{and} \qquad \bPsi^k_{\rho} = \sum_{i=1}^{k} \sum_{j=i}^{k} \rho^{j-i}\tilde{\bPsi}_{i,j}.
\end{equation}
Then,
{\allowdisplaybreaks
\begin{align}
    \left\| \bPsi_{\rho}-\bPsi_{\rho}^k \right\|
    & = \left\| \sum_{i=1}^{\infty} \sum_{j=i}^{\infty} \rho^{j-i}\tilde{\bPsi}_{i,j}-\sum_{i=1}^{k} \sum_{j=i}^{k} \rho^{j-i}\tilde{\bPsi}_{i,j} \right\| \nonumber \\
    & = \left\| \sum_{i=k+1}^{\infty} \sum_{j=i}^{\infty} \rho^{j-i}\tilde{\bPsi}_{i,j}+\sum_{i=1}^{k} \sum_{j=k+1}^{\infty} \rho^{j-i}\tilde{\bPsi}_{i,j} \right\| \nonumber \\
    & \leq  \sum_{i=k+1}^{\infty} \sum_{j=i}^{\infty} \rho^{j-i}\left\|\tilde{\bPsi}_{i,j}\right\| +\sum_{i=1}^{k} \sum_{j=k+1}^{\infty} \rho^{j-i}\left\|\tilde{\bPsi}_{i,j} \right\| \nonumber \\
    & = \sum_{i=k+1}^{\infty} \sum_{j=i}^{\infty} \rho^{j-i}\left\|\bPsi_{i,j}\right\| +\sum_{i=1}^{k} \sum_{j=k+1}^{\infty} \rho^{j-i}\left\|\bPsi_{i,j} \right\| \nonumber \\
    & = \sum_{i=1}^{\infty} \sum_{j=i}^{\infty} \rho^{j-i}\left\|\bPsi_{i,j}\right\| - \sum_{i=1}^{k} \sum_{j=1}^{k} \rho^{j-i}\left\|\bPsi_{i,j} \right\| \underset{k \to \infty}{\longrightarrow} 0,
\end{align}}
where in the third step we used the fact that
\begin{equation}
    \sum_{i=1}^{\infty} \sum_{j=i}^{\infty} \rho^{j-i}\left\|\tilde{\bPsi}_{i,j}\right\| = \sum_{i=1}^{\infty} \sum_{j=i}^{\infty} \rho^{j-i}\left\|\bPsi_{i,j}\right\| 
\end{equation}
is bounded (see \eqref{eq:bound block sum}). Therefore, $ \bPsi_{\rho}^k \overset{\| \cdot \|}{\longrightarrow} \bPsi_{\rho} $ as $ k \to \infty $, and thus $ \bPsi_{\rho} $ is compact. This means that the non-zero elements in the spectrum of $ \bPsi_{\rho} $ are comprised of its eigenvalues only (point spectrum).

In the following, we use a known result about the convergence of the spectrum of finite rank approximations. 
\begin{lemma}[{\protect\citet[Cp.~XI.9 Lemma~5]{dunford1964linear}}]\label{lemma:Eigenvalues of compact operator}
Let $ \{\TT_k\}$ and $\TT $ be compact operators, such that $ \TT_k \overset{\| \cdot \|}{\longrightarrow} \TT $. Let $ \lambda_m(\TT) $ be an enumeration of the non-zero eigenvalues of $\TT$, each repeated according to its multiplicity. Then there exist enumerations $ \lambda_m(\TT_k) $ of the non-zero eigenvalues of $\{\TT_k\}$, with the repetitions according to multiplicity, such that
\begin{equation}
    \lim_{k \to \infty}\lambda_m(\TT_k) = \lambda_m(\TT), \qquad m \geq 1,
\end{equation}
where the limit is uniform in $ m $.
\end{lemma}

Let $ \sigma(\cdot) $ denote the spectrum of an operator. Here, each $ \bPsi_{\rho}^k $, when restricted to its square support, is a block upper triangular matrix. Hence, its spectrum\footnote{Without the zero eigenvalue.} is given by the union of the eigenvalues of the blocks in the diagonal. Namely,
\begin{equation}
    \sigma \left( \bPsi_{\rho}^k \right) = \bigcup_{j = 1}^k \sigma \left(\bPsi_{j,j}\right).
\end{equation}
Thus, according to Lemma~\ref{lemma:Eigenvalues of compact operator} we have that the non-zero spectrum of $ \bPsi_{\rho} $ is
\begin{equation}\label{eq:operator spectrum}
    \sigma\left(\bPsi_{\rho}\right) \backslash \{ 0 \} = \lim_{k \to \infty}  \sigma \left( \bPsi_{\rho}^k \right) \backslash \{ 0 \} = \bigcup_{k = 1}^{\infty} \sigma \left(\bPsi_{k,k}\right) \backslash \{ 0 \}.
\end{equation}

Now that we have the spectrum of $ \bPsi_{\rho} $ we turn to show that under the condition of Thm.~\ref{Thm:Sufficient condition for SGD} in \eqref{eq:Sufficient condition}, the spectral radius $ r(\bPsi_{\rho} ) $ is less than one. Due to \eqref{eq:operator spectrum}, it is sufficient to show that for all $ k \in \N $ we have $ r(\bPsi_{k,k}) \leq c < 1 $, for some constant $ c \in (0,1) $. Recall that 
\begin{equation}
    \bPsi_{k,k} = \E\left[ \left( \mD \SGDtransform_{\batch_t}(\params^*)  \right)^{\otimes k} \right],
\end{equation}
where $\mD \SGDtransform_{\batch_t}(\params^*) = \IdentityMat - \eta \nabla^2 \stochasticLoss_{\batch_t}(\params^*) $ is symmetric. Therefore, $ \bPsi_{k,k} $ is also symmetric, and we have that $ r(\bPsi_{k,k}) = \| \bPsi_{k,k} \|$. Thus, using Jensen's inequality
\begin{align}
    r\left(\bPsi_{k,k}\right) 
    & = \left\| \bPsi_{k,k} \right\| \nonumber \\
    & = \left\| \E\left[ \left( \mD \SGDtransform_{\batch_t}(\params^*)  \right)^{\otimes k} \right] \right\|  \nonumber \\
    & \leq \E\left[ \left\| \left( \mD \SGDtransform_{\batch_t}(\params^*)  \right)^{\otimes k}\right\|  \right]  \nonumber \\
    & = \E\left[ \left\| \mD \SGDtransform_{\batch_t}(\params^*)  \right\|^{k}  \right].
\end{align}
Note that under the conditions of Thm.~\ref{Thm:Sufficient condition for SGD} we have $ \left\| \mD \SGDtransform_{\batch_t}(\params^*)  \right\| \leq  1-\varepsilon $ w.p. $1$ for some $ \varepsilon \in (0,1) $ (see \eqref{eq:singular value bound}, and the discussion above it). Therefore, 
\begin{equation}
     r\left(\bPsi_{k,k}\right) \leq \E\left[ \left\| \mD \SGDtransform_{\batch_t}(\params^*)  \right\|^k  \right] \leq (1-\varepsilon)^k.
\end{equation}
Overall,
\begin{equation}
    r\left( \bPsi_{\rho} \right) = \sup_{k \in \N} \ r\left(\bPsi_{k,k}\right) \leq \sup_{k \in \N} \ (1-\varepsilon)^k = 1-\varepsilon<1.
\end{equation}

\section{Proof of Theorem~\protect\ref{thm:boundness test}}\label{Appendix:BoundnessSufficientCondition}
Let $ \TT $ be an operator defined on $ \ell_2 $ space. Assume that $\TT $ consists of blocks $ \{ \TT_{i,j} \}$, such that $ \forall i,j \; \TT_{i,j} \in \R^{d_i \times d_j}$ where $ \{ d_i \}_{i=1}^\infty $ is a given sequence. Additionally, assume that 
\begin{equation}
    \forall j \in \N \qquad \sum_{i = 1}^\infty \|\TT_{i,j}\| \leq \alpha \qquad \text{and} \qquad \forall i \in \N \qquad \sum_{j = 1}^\infty \|\TT_{i,j}\| \leq \beta.
\end{equation}
Furthermore, for any $ \uu \in \ell_2$, denote by $ \uu_i $ its $ i$th segment, such that $ \uu_i \in \R^{d_i} $. Then we have
{\allowdisplaybreaks
\begin{align}
    \| \TT \uu \|^2 
    & = \sum_{i = 1}^\infty \left\| \sum_{j=1}^{\infty} \TT_{i,j} \uu_j \right\|^2 \nonumber \\
    & \leq \sum_{i = 1}^\infty \left( \sum_{j=1}^{\infty} \| \TT_{i,j} \| \| \uu_j \| \right)^2 \nonumber \\
    & = \sum_{i = 1}^\infty \left( \sum_{j=1}^{\infty} \sqrt{\| \TT_{i,j} \|} \sqrt{ \| \TT_{i,j} \| } \| \uu_j \| \right)^2 \nonumber \\
    & \leq \sum_{i = 1}^\infty \left( \sum_{j=1}^{\infty} \| \TT_{i,j} \|  \right) \left( \sum_{j=1}^{\infty} \| \TT_{i,j} \| \| \uu_j \|^2 \right) \nonumber \\
    & \leq \sum_{i = 1}^\infty \beta \left( \sum_{j=1}^{\infty} \| \TT_{i,j} \| \| \uu_j \|^2 \right) \nonumber \\
    & = \beta \sum_{j=1}^{\infty} \| \uu_j \|^2  \sum_{i = 1}^\infty   \| \TT_{i,j} \|  \nonumber \\
    & \leq  \beta \sum_{j=1}^{\infty} \| \uu_j \|^2  \alpha \nonumber \\
    & = \alpha \beta \| \uu \|^2,
\end{align}}
where in the second step we used the triangle inequality, the fourth step is due to the Cauchy-Schwarz inequality, and in the sixth step we used the fact that all summands are non-negative, and therefore we can change summation order.

\end{document}